\documentclass{article}

\usepackage{microtype}
\usepackage{graphicx}
\usepackage{subfigure}
\usepackage{booktabs} % for professional tables

\usepackage{algorithm2e}
\usepackage{amssymb, url, amsmath, relsize}

\usepackage{mathrsfs, enumerate}
\graphicspath{ {figures/} }
\usepackage{mathtools}
\usepackage{multicol}
\usepackage{wrapfig}
\usepackage{makecell}
\usepackage{xcolor}
\usepackage{amsmath,amsfonts,bm}

\def\eqref#1{\usepackage{xcolor}
equation~\ref{#1}}
\def\1{\bm{1}}

\DeclareMathAlphabet{\mathsfit}{\encodingdefault}{\sfdefault}{m}{sl}
\SetMathAlphabet{\mathsfit}{bold}{\encodingdefault}{\sfdefault}{bx}{n}

    \usepackage[preprint]{neurips_2021}

\usepackage[utf8]{inputenc} % allow utf-8 input
\usepackage[T1]{fontenc}    % use 8-bit T1 fonts
\usepackage{url}            % simple URL typesetting
\usepackage{booktabs}       % professional-quality tables
\usepackage{amsfonts}       % blackboard math symbols
\usepackage{nicefrac}       % compact symbols for 1/2, etc.
\usepackage{microtype}      % microtypography
\usepackage{xcolor}         % colors

\usepackage{algorithm}
\usepackage{algorithmic}
\usepackage{xfrac}
\usepackage{pgfplots}
\usepackage[export]{adjustbox}

\newtheorem{theorem}{Theorem}[section]

\newtheorem{lemma}[theorem]{Lemma}

\title{CobBO: Coordinate Backoff Bayesian Optimization with Two-Stage Kernels}

\author{%
  Jian Tan\thanks{Equal contribution} \\
  Alibaba Group \\ Sunnyvale, California, USA \\
   \And
   Niv Nayman$^*$ \\
   Alibaba Group \\ Tel Aviv, Israel \\
   \And
   Mengchang Wang\\
   Alibaba Group \\ Hangzhou, Zhejiang, China \\
   \AND
   \texttt{\{j.tan, niv.nayman, mengchang.wmc\}@alibaba-inc.com} \\
}

\begin{document}

\maketitle

\begin{abstract}
Bayesian optimization is a popular method for optimizing expensive black-box functions. 
Yet it oftentimes struggles in high dimensions where the computation could be prohibitively heavy. 
While a complex kernel with many length scales are prone to overfitting and expensive to train, a simple kernel with too few length scales cannot effectively capture the variations of the high dimensional function in different directions.
To alleviate this problem, we introduce Coordinate backoff Bayesian Optimization (CobBO) with two-stage kernels.  
During each round, the first stage uses a simple coarse kernel that sacrifices the approximation accuracy for computational efficiency. It captures the global landscape by purposely smoothing away local fluctuations. Then, in the second stage of the same round, past observed points in the full space are projected to the selected subspace to form virtual points. 
These virtual points, along with the means and variances of their unknown function values estimated using the simple kernel of the first stage, are fitted to a more sophisticated kernel model in the second stage. Within the selected low dimensional subspace, the computational cost of conducting Bayesian optimization therein becomes affordable. To further enhance the performance, a sequence of consecutive observations in the same subspace are collected, which
can effectively refine the approximation of the function. 
This refinement lasts until a stopping rule is met determining when to back off from a certain subspace and switch to another.
This decoupling significantly reduces the computational burden in high dimensions,
which fully leverages the observations in the whole space rather than only relying on observations in each coordinate subspace.
Extensive evaluations show that CobBO finds solutions comparable to or better than other state-of-the-art methods for dimensions ranging from tens to hundreds, while reducing both the trial complexity and computational costs.
\end{abstract}
\section{Introduction}\label{s:intro}
%\vspace{-1.5mm}
Bayesian optimization (BO) has emerged as an effective zero-order paradigm for optimizing expensive black-box functions. 
It has been widely used in various real applications, e.g.,  parameter tuning for recommendation systems, automatic database configuration tuning, and simulation-based optimization.
%The entire sequence of iterations rely only on the function values of the queried points without information on their derivatives. 
%Of significant interest, one cares not only the trial complexity, i.e., the number of queried points,
% but also the time complexity, i.e., the total execution time. 
% The latter consists of the time spent on suggesting the points 
% and on evaluating their corresponding function values. 

Though highly competitive in low dimensions (e.g., the dimension $D\leq 20$~\cite{frazier2018}),  
Bayesian optimization based on Gaussian Process (GP) regression has obstacles %that impede its effectiveness, 
in high dimensions. 
 %To overcome the hurdles of applying Bayesian optimization, inevitably one 
%needs to reduce not only the trial complexity, i.e., the number of queried data points,
 %but also the time complexity, i.e., the total execution time. 
 %The latter consists of the time spent on suggesting points 
 %and on evaluating their corresponding function values. 
% \begin{enumerate}
%  \item 
\\ 
\textbf{Curse of dimensionality}:  As a sample efficient method, Bayesian optimization often suffers from high dimensions. Fitting the GP model (estimating the parameters, e.g., length scales~\cite{turbo2019})
%computing the Gaussian process posterior 
and optimizing the acquisition function all incur large computational costs in high dimensions. It also results in statistical insufficiency of exploration~\cite{josip2013,zi2017}.  
As the GP regression’s error grows with dimensions~\cite{bull2011}, more samples are required to balance that in high dimensions, which
could cubically increase the computational costs in the worst case~\cite{mutny2018}. Undesirably, the computation times, especially for model fitting and acquisition function optimization, 
%of a vanilla BO algorithm in high dimensions 
could be even far longer than the required time for evaluating the function values in high dimensions, which significantly limits the application.  
\\
% \vspace{-3mm} \\
% \textbf{Approximation accuracy}: 
 \textbf{Multiple length scales}: 
 %GP regression assumes a class of random functions in a probability space
 %as surrogates that iteratively yield posterior distributions by conditioning on the queried points. 
 %When suggesting new query points,
 %for complex functions with numerous local optima and saddle points due to local fluctuations, always exactly using the values on the queried points as the conditional events may mismatch the function's local landscape by overemphasizing the approximation accuracy of the global landscape. 
 The smoothness of the regression is determined by the specified kernel and the corresponding length scales, where the latter can be viewed as the measuring units along different coordinates. 
 %For many real world problems 
 %The local fluctuations and the global landscape of the function jointly impact the approximation accuracy, 
 The landscapes of the function on the global full space and on different local coordinate subspaces can vary significantly as BO tries to approximate all of them in each iteration using a family of Gaussian functions. %from the global full space. 
 %Thus, there is no single set of length scales that fits all. 
 Thus, a single kernel with a fixed set of length scales cannot effectively fit all. 
 %For properly capturing the local fluctuations of a function near a local optimum, a short length scale is required.

%  \begin{algorithm}[th]
% 	\caption{High level description of CobBO}
% 	\label{alg:high_level}
% \begin{algorithmic}[1] 
% \FOR{each round $r$}
%     \STATE \textbf{Stage 1}: 
%     \STATE $F_1 \leftarrow$ conduct GP regression using all of the observed data points on the full space and a simple kernel that is easy to compute
%     \STATE Select a subspace $\Omega_r$, project the observed points onto $\Omega_r$ to obtain a new set $\mathcal{X}_r$ of ``virtual points'' and estimate by $F_1$ their function values if unknown
%     \STATE \textbf{Stage 2}: 
%     \WHILE{Stopping rule is not met}
%     \STATE $F_2 \leftarrow$ conduct GP regression on the subspace $\Omega_r$ using $\mathcal{X}_r$ and a sophisticated kernel
%     \STATE Conduct BO using $F_2$ on $\Omega_r$ and suggest the next query point within $\Omega_r$
%     \STATE Evaluate the function value of the new point and add it to $\mathcal{X}_r$
%     \ENDWHILE
% \ENDFOR 
%   \STATE return the best observed data point and its function value
% \end{algorithmic}
% \end{algorithm}

  \begin{algorithm}[th]
	\caption{High level description of CobBO}
	\label{alg:high_level}
\begin{algorithmic}[1] 
\FOR{each round $r$}
    \STATE \textbf{Stage 1}: 
    \STATE GP regression using a computation-efficient kernel $K_1$ on all of the observed data points from the full space $\Omega$
    \STATE Select a subspace $\Omega_r$, construct  ``virtual points'' and estimate the means (and optional variances) of their function values using $K_1$
    \STATE \textbf{Stage 2}: 
    \REPEAT
    \STATE BO on the same subspace $\Omega_r$ with a sophisticated and maybe time-consuming kernel $K_2$ using both the ``virtual points'' and truly observed ones on $\Omega_r$
    \UNTIL{backoff stopping rule is met}
\ENDFOR 
  \STATE \textbf{return} the best observed data point
\end{algorithmic}
\end{algorithm}

% \\ \vspace{-3mm} \\
%\\ \vspace{-3mm} \\
%\textbf{Stagnation at local optima}:
%It is known that Bayesian optimization could stagnate at local %optima~\cite{qin2017,bull2011,snoek2012}.

 \begin{figure*}
        \centering
	\includegraphics[width=0.99\linewidth,height=!]{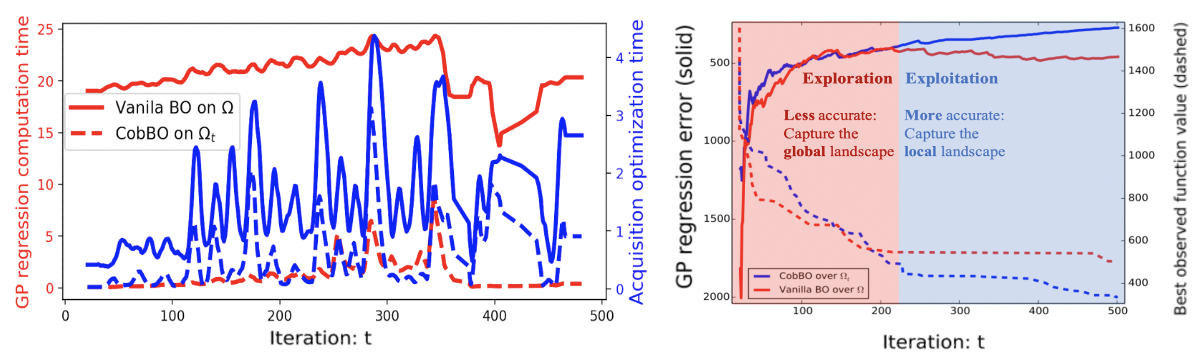}
  \vspace{-3mm}
	\caption{Minimize the fluctuated Rastrigin function on $[-5, 10]^{50}$ with $20$ initial samples. [Left] Computation times for training the GP regression model and maximizing the acquisition function at each iteration. CobBO significantly reduces the execution time compared with a vanilla BO, e.g. $\times 13$ faster in this case. [Right] The average error between the GP predictions before making queries and the true function values at the queried points
	(solid curves, the higher the better) and the best observed function value (dashed curves, the lower the better) at iteration~$t$. 
	During each round, CobBO captures the global landscape less accurately using the RBF kernel at the first stage, and then explores selected subspaces $\Omega_t$ more accurately using the Matern kernel at the second stage. This eventually better exploits the promising subspaces.}
	\label{fig:motivation}
	%\vspace{-6mm}
\end{figure*}

To alleviate, we introduce two-stage kernels and design coordinate backoff Bayesian optimization (CobBO), as illustrated in Algorithm~\ref{alg:high_level}. 
The first stage uses a computation-efficient kernel for capturing the global landscape in the full space. For example, one can use a simple and coarse kernel to purposely smooth away local fluctuations, 
%which is also cheap to compute, 
e.g., RBF~\cite{rbf}, where efficient algorithms in $O(N\log N)$ for $N$ observations have been well studied~\cite{Gumerov07}. 
%Specifically, using a `multiquadric' kernel with length scales approximating the average distance between points, CobBO can efficiently fit the model in the full space. Other efficient methods also exist; see Section~\ref{}. 
In contrast, the second stage utilizes a more sophisticated kernel, e.g., the Automatic Relevance Determination (ARD) Mat\'{e}rn 5/2~\cite{matern}, that 
 %fits a model to 
 learns the varying length scales to properly capture the local fluctuations in the selected subspaces.
 
 To bridge the two stages, CobBO introduces \textit{virtual points} in the first stage by estimating the values of the unobserved points from the queried ones projected to selected promising subspaces. Then it conducts BO by conditioning on both \textit{virtual points} and real observations that reside in the selected subspace, 
 %This is different from a common approach that is directly conditioning on the queried points.
by which the information outside the subspace can also be effectively utilized.
 This introduced decoupling allows us to apply two different kernels on the global space and the subspace, respectively. It can dramatically reduce both the model fitting time in the full space and the acquisition function optimization time in the subspace, as shown in Fig~\ref{fig:motivation}.  

This method can be viewed as a variant of block coordinate ascent tailored to Bayesian optimization. 
%by applying backoff stopping rules for switching coordinate blocks.  
While similar works exist~\cite{dropoutbo,moriconi2020,Oliveira2018}, CobBO differs
by introducing the two-stage kernels and
the following features: 
\begin{enumerate}
% \vspace{-1mm}
\itemsep0em
% \item A coordinate subspace requires a sufficient number of query points acting as the conditional events for the GP regression. CobBO leverages all observations in the whole space by interpolating the values of queried points projected to selected promising subspaces, rather than simply starting from scratch in each subspace. 
% The two-stage kernel Gaussian process regressions fully leverage the observations in the whole space rather than only relying on observations in each coordinate subspace.
% \item A coordinate subspace requires a sufficient amount of query points acting as the conditional events for the GP regression.  Without enough points and corresponding values, the function landscape within a subspace cannot be well-characterized~\cite{bull2011}. 
\item To refine the approximation in a subspace and also reduce the computation time,  the second stage of CobBO relies on a sequence of observations determined by a stopping rule that backs off from a certain subspace and switches to another one.
When consecutively querying data points in the same subspace, CobBO does not need to conduct the first-stage, including the model fitting and the GP regression, in the full space. % which is far more efficient.   Notably, 
In addition, in the second-stage on a low dimensional subspace, both computing the Gaussian process posterior and optimizing the acquisition function can be efficiently conducted, moderating the curse of dimensionality. 
However, querying a certain subspace %, under some trial budget, 
comes at the expense of exploring other coordinate blocks. Yet prematurely shifting to different subspaces does not fully exploit the full potential of a given subspace. Hence determining the number of consecutive function queries within a subspace makes a trade-off between exploration and exploitation. 
%CobBO uses a stopping rule in each subspace to switch the selected coordinates. 
\item Selecting a block of coordinates requires determining the block size as well as the coordinates therein. 
%The coordinate subsets are selected by using
%a multiplicative weights update method~\cite{sanjeev12} to the preference probability associated with each coordinate.
CobBO selects the coordinate subsets by
a multiplicative weights update method~\cite{sanjeev12} to the preference probability associated with each coordinate. %uses a multiplicative weights update method 
Thus, it samples more promising subspaces with higher probabilities. 
\end{enumerate}

%To this end, we design CobBO to address these challenges.
% The coordinate subsets are selected by using
% a multiplicative weights update method~\cite{sanjeev12} to the preference probability associated with each coordinate.
% The trade-off between exploration and exploitation is balanced by the subspace selection and the switching,  governed by the preference probability and the backoff stopping rule, respectively.
% In addition, differently from~\cite{luigi2017,javier2016,McLeod2018OptimizationFA,turbo2019}, CobBO dynamically forms trust regions on two time scales to further tune this trade-off.

Through comprehensive evaluations, CobBO demonstrates appealing performance for dimensions ranging from tens to hundreds.
It obtains comparable or better solutions with fewer queries, in comparison with the state-of-the-art methods, for most of the problems tested in Section~\ref{s:exp}. 

% \vspace{-0.0mm}
\section{Related work} \label{sec:related_work}
% \vspace{-0.0mm}
%Given the large body of work on Bayesian optimization~\cite{frazier2018,Brochu2010,shahriari2016}, 
%we summarize the most relevant ones on high dimensionality. 
%in the following topics: high dimensionality, 
%trust regions, 
%hybrid methods and batch sampling.
%In Section~\ref{s:exp}, we compare CobBO with selected and competitive algorithms. 
% For an overview of Bayesian optimization see.
%\textbf{High dimensionality:}
%To apply Bayesian optimization in high dimensions, 
Certain assumptions
are often imposed on the latent structure in high dimensions. 
Typical assumptions include low dimensional embedding and additive structures. Their advantages manifest on problems with a low dimension or a low effective dimension. 
However, these assumptions do not necessarily hold in practice, e.g., for non-separable functions without redundant dimensions.

\noindent \textbf{Low dimensional embedding:} 
The function $f$ is assumed to have a low effective dimension~\cite{kushner1964,hemant2014}, e.g.,  $f(x) = g(\Phi x)$ for a function $g(\cdot)$ and a matrix $\Phi$ of $d\times D, d<<D$. It essentially assumes that $f(x)$ does not change along certain directions. 
A variety of methods have been developed, including
random embedding~\cite{josip2013,ziyuw2016,chaudhuri2019,binois2019,letham2020},
Hashing-enhanced Subspace BO (HeSBO)~\cite{chaudhuri2019}, and Mahalanobis kernel %for linear embeddings 
ALEBO~\cite{letham2020}.
%DROPOUT~\cite{dropoutbo} and LineBO~\cite{linebo}.
%~\cite{ziyu2013,josip2013,ziyuw2016,chuliang2016,chaudhuri2019,miao2019,binois2019,letham2020},
%(e.g., REMBO~\cite{ziyuw2016}),  
%low-rank matrix recovery~\cite{josip2013,hemant2014},
%and learning subspaces by derivative information~\cite{josip2013,eriksson2018}.
%In contrast to existing work on subspace selections,
%LineBO which receives a special treatment in Appendix~\ref{ss:linebo}, 
%CobBO efficiently leverages all the observations in the whole space using the two-stage kernels and the stopping rule in each subspace for consecutive observations. 
%rather than only relying on limited observations in each coordinate subspace. 
%While the simple first-stage kernel involves all the coordinates in the computation, the second stage conducts both the GP regression and the learning of the length scales of the more sophisticated kernel in the lower dimensional subspaces. 
%DROPOUT selects the active coordinates at every iteration and fill in the remaining coordinates using some heuristic strategy. CobBO uses a simple kernel applied on the full space to estimate the function values of newly added “virtual points” on the subspace from all past data.
%rather than simply starting from scratch in each subspace. 
%The two-stage kernel Gaussian process regressions fully leverage the observations in the whole space rather than only relying on observations in each coordinate subspace. 
Since not all the real-world problems fit the low dimensional embedding structure, CobBO is designed to optimize functions without redundant dimensions. It exploits the subspace structure, independent of the dimensions.
Though the embedding-based algorithms and CobBO are based on different assumptions, REMBO~\cite{ziyuw2016} and ALEBO~\cite{letham2020} are compared with CobBO in Section~\ref{ss:alebo}. 
%As a result, it shows great performance in both high and low dimensions, 
%different from some algorithms that are more suitable for low dimensions, e.g., BADS~\cite{luigi2017}. 

%SI-BO~\cite{josip2013}

\noindent \textbf{Additive structure}:
%Sparse Gaussian processes~\cite{mitchell2016},
A decomposition assumption is often made by $f(x) = \sum_{i=1}^{k}f^{(i)}\left(x_{i} \right)$, with $x_i$
defined over low-dimensional components.  In this case, the effective dimensionality of the model is
the largest dimension among all additive groups~\cite{mutny2018}, which is usually small.  
The Gaussian process is structured as an additive model~\cite{elad2013,kandasamy2015}.
%e.g.,  projected-additive functions~\cite{chuliang2016}, ensemble Bayesian optimization (EBO)~\cite{wang18aistats}, latent additive structural kernel learning (HDBBO)~\cite{zi2017} and group additive models~\cite{kandasamy2015,chuliang2016}. 
%Though this method effectively reduces
%the time complexity, it has been reported that the accuracy is often slightly lower than a full GP~\cite{elad2013}.  
However, learning the unknown structure incurs a considerable computational cost~\cite{chaudhuri2019}, and is not always applicable for non-separable functions.
%for which CobBO can still be applied. % in these scenarios.   %As a variant of the block coordinate ascent method, 
%CobBO can be applied for non-separable functions.  

% \vspace{-1mm}
% \emph{Kernel methods:}
% Various kernels have been used for resolving the difficulties in high dimensions, 
% e.g., 
% a hierarchical Gaussian process model~\cite{chen2019hierarchical}, a cylindrical kernel~\cite{bock2018} and a compositional kernel~\cite{david2013}.
% CobBO can be integrated with other sophisticated methods~\cite{snoek2012, david2013, bock2018,chen2019hierarchical,marchuk1975,jones1998,srinivas2010,frazier2008,scott2011,ziw2017}, e.g., ATPE/TPE~\cite{TPE2011,ATPE}
% and SMAC~\cite{HutHooLey11-smac}.

% \vspace{-0.8mm}
% \textbf{Trust regions and space partitions:}
% \vspace{-0mm}
\noindent \textbf{Trust regions and subspaces:}
Trust region BO has been proven effective for high-dimensional problems.
%A typical pattern is to alternate between global and local search regions. 
Within the local trust regions, many efficient methods have been applied, e.g.,  local Gaussian models (TurBO~\cite{turbo2019}),  adaptive search on a mesh grid (BADS~\cite{luigi2017}) or quasi-Newton local optimization (BLOSSOM~\cite{McLeod2018OptimizationFA}). 
TurBO~\cite{turbo2019} uses Thompson sampling to allocate samples across multiple regions.
A related method is to use space partitions, e.g., LA-MCTS\cite{Wang2020LearningSS} on a Monte Carlo tree search algorithm to learn efficient partitions. 
CobBO differs by selecting low dimensional subspaces. 
 A closely related work is LineBO~\cite{linebo}, which also significantly reduces the acquisition function optimization time by restricting on one-dimensional subspaces. However, as it uses a single kernel, it does not address the computational issue of the GP regression in the full space. See a comparison in Section~\ref{ss:linebo}.

\section{Algorithm}\label{sec:algorithm}
% \vspace{-0.0mm}
Without loss of generality, suppose that the goal is to solve 
%a problem
$x^{\ast} = \textrm{argmax}_{x\in \Omega} f(x)$
for a black-box function $f: {\Omega} \to \mathbb{R}$.
The domain is normalized ${\Omega} = [0,1]^D$ with the coordinates indexed by $I=\{1,2, \cdots, D\}$. For a sequence of points $\mathcal{X}_t = \{x_1,  x_2, \cdots, x_t\}$ with $t$ indexing the most recent iteration, we observe $\mathcal{H}_t=\left\{ \left(x_i, y_i=f(x_i)\right) \right\}_{i=1}^t$. A subset $C_t \subseteq I$ of the coordinates is selected, forming a subspace $\Omega_t \subseteq \Omega$.
%CobBO uses Bayesian optimization, hence essentially sequential. 
%\niv{Why are BO "essentially sequential" ? What about batch versions of BO ?} 
%While CobBO involves several hyperparameters, extensive experiments demonstrate CobBO's robustness to those as it achieves great performance across the many tasks in Section~\ref{s:exp} using the same default configuration.
%\niv{The above sentences are out of context in the middle of the settings description}
%Due to selecting random subspaces, it can be easily paralleled by batch sampling. %as explained in Section~\ref{ss:batch}.  
%To simplify the presentation, we focus on the sequential mode in this section. 

 GP regression assumes a class of random functions in a probability space
 as surrogates that iteratively yield posterior distributions by conditioning on the queried points.
For iteration $t$, instead of computing the Gaussian process posterior distribution
 $\left\{ \hat{f}(x) \middle| \mathcal{H}_t = \left\{ (x_i, y_i)\right\}_{i=1}^t, x \in {\Omega} \right\}$ by conditioning on the observations
  $y_i=f(x_i)$ at queried points~$x_i$ in the full space $\Omega \subset \mathbb{R}^D$ for $i=1,\dots,t$, we change the conditional events, and consider
  \begin{align*}
  \left\{ \hat{f}(x) \middle| R\left( P_{\Omega_t} (x_1,\dots,x_t), \mathcal{H}_t\right), x \in {\Omega}_t, {\Omega}_t \subset \Omega \right\}
  \end{align*}
  for a projection function $P_{\Omega_t}(\cdot)$ to a random subspace ${\Omega}_t$ and an estimation function $R(\cdot, \cdot)$.
  %e.g., using a RBF approximation without learnable parameters~\cite{rbf} as the simple kernel for the first stage. 
The projection $P_{\Omega_t}(\cdot)$ maps the queried points to virtual points on a subspace~${\Omega}_t$ of a lower dimension~\cite{ali2008}. 
The function $R(\cdot, \cdot)$ estimates means and variances of the objective values at the virtual points using $\mathcal{H}_t$. The second stage within the subspace $\Omega_t$ 
uses a more sophisticated kernel that otherwise would be expensive to learn the parameters and incur high computational cost in high dimensions. 
%This decoupling significantly reduces the computational burden.

As a variant of coordinate ascent, the subspace $\Omega_t$ contains a pivot point $V_t$, which presumably is the maximum point $x^M_t = \textrm{argmax}_{x\in \mathcal{X}_t}f(x)$ with $M_t = f\left(x^M_t\right)$. CobBO may set $V_t$ different from $x^M_t$ to escape local optima, as is a well-known issue for coordinate ascent. 
%$M_t = \max_{x\in \mathcal{X}_t} f(x)$.
Then, BO is conducted within $\Omega_t$
%to maximize an acquisition function based on $f(x)$ 
while fixing all the other coordinates $C^c_t= I\setminus C_t$,
i.e.,  the complement of $C_t$.
% However, when the number $q_t$ of consecutive queries at iteration $t$ that fail to improve over $M_{t-1}$ becomes larger than a threshold $\Theta$,
% %(e.g., $\Theta=70$), 
% we decrease the observed function value at $V_{t-1}$ and set $V_{t}$ as a selected sub-optimal random point in $\mathcal{X}_t$ in order to escape a trapped local maxima. Specifically, we randomly sample a few points in $\mathcal{X}_t$ with their values at the top half and pick the one furthest away from $V_{t-1}$. 

% \input{algorithm/algorithm_1_cobbo}

\begin{algorithm}[th]
	\caption{Detailed description of CobBO(f, $\tau$, T)}
	\label{alg:top}
\begin{algorithmic}[1] 
\STATE $\mathcal{H}_{\tau} \leftarrow$ sample $t=\tau$ initial points and evaluate their values
\STATE $V_\tau, M_\tau \leftarrow$ the maximal point in $\mathcal{H}_{\tau}$
\STATE $q_{\tau} \leftarrow 0$, number of consecutive failed queries
\STATE $\pi_{\tau} \leftarrow$ uniform preference for coordinates
\WHILE{$t < T$}
\STATE \textbf{Stage 1}: use a computation-efficient kernel $K_1$ for the estimation function $R(\cdot, \cdot)$	
   \STATE   $\Omega_t \leftarrow$ take a new subspace over a coordinate block $C_t$, satisfying $V_t\in \Omega_t$
   \STATE $\hat{\mathcal{X}}_{t} \leftarrow$ $P_{\Omega_t}(\mathcal{X}_{t})$ 
    //form virtual points %$\mathcal{X}_{t}$
	                   %onto $\Omega_t$ to obtain a set of virtual points (Eq.~\ref{eq:projection})\Big{]}
    \STATE $\hat{\mathcal{H}}_{t} \leftarrow$ $R\left( \hat{\mathcal{X}}_t, \mathcal{H}_t\right)$ //compute the means (and optional variances) of the virtual points by $K_1$ 
	                   % \Big{[}Smooth function values on $\hat{\mathcal{X}}_{t}$ by interpolation using $\mathcal{H}_t$\Big{]} 
	                   
\STATE \textbf{Stage 2}: use a sophisticated kernel $K_2$ for BO within $\Omega_t$ for consistent queries
\REPEAT
\STATE $p\left[ \hat{f}_{\Omega_t}(x) | \hat{\mathcal{H}}_{t} \right] \leftarrow$ compute the posterior distribution of the Gaussian process in $\Omega_t$ conditional on $\hat{\mathcal{H}}_{t}$ by $K_2$; note that the points in $\hat{\mathcal{H}}_{t}$ could have non-zero variances
\STATE ${x}_{t+1} \leftarrow \textrm{argmax}_{x\in \Omega_t} Q_{ \hat{f} \sim p(\hat{f}|\hat{\mathcal{H}}_t)}(x  | \hat{\mathcal{H}}_t)$ //keep using $K_2$
        %  \Big{[} Suggest the next query in $\Omega_t$ (Section~\ref{sec:algorithm})\Big{]}
\STATE $y_{t+1} = f(x_{t+1})$
% \STATE $y_{t+1} \leftarrow$ Evaluate the black-box function $y_{t+1} = f(x_{t+1})$
\IF{$y_{t+1} > M_t$}
\STATE  $V_{t+1} \leftarrow x_{t+1}$, $M_{t+1} \leftarrow y_{t+1}$, $q_{t+1} \leftarrow 0$
\ELSE
    \STATE $V_{t+1} \leftarrow V_t$, $M_{t+1} \leftarrow M_t$, $q_{t+1} \leftarrow q_t + 1$
\ENDIF 
% \STATE $\pi_{t+1} \leftarrow $ Update $\pi_{t}$ by a multiplicative weights update method (Eq.~\ref{eq:multiplicative_update})
\STATE Update $\pi_{t}$ according to Eq.~(\ref{eq:multiplicative_update})
\STATE $\Omega_{t+1} \leftarrow \Omega_{t}$ //remain in the same subspace for the next query
\STATE $\mathcal{H}_{t+1} \leftarrow \mathcal{H}_{t} \bigcup \{(x_{t+1}, y_{t+1})\}$, $\mathcal{X}_{t+1} \leftarrow \mathcal{X}_{t} \bigcup \{x_{t+1}\}$, $t\leftarrow t+1$
\UNTIL{switch to a different subspace by the backoff stopping rule}
\ENDWHILE
\end{algorithmic}
\end{algorithm}

For BO in $\Omega_t$, we use Gaussian processes as the random surrogates $\hat{f}=\hat{f}_{\Omega_t}(x)$ to describe the Bayesian statistics of $f(x)$ for $x \in \Omega_t$.  
At each iteration, the next query point is by solving
\begin{align*}
% 	 \vspace{-1.5mm}
  x_{t+1} = \textrm{argmax}_{x\in \Omega_t, V_t \in \Omega_t} Q_{ \hat{f}_{\Omega_t}(x) \sim p(\hat{f}|\mathcal{H}_t)}(x | \mathcal{H}_t),
	 \vspace{-1.5mm}
\end{align*}
where the acquisition function $Q(x | \mathcal{H}_t)$ incorporates the posterior distribution of the Gaussian processes $p(\hat{f}|\mathcal{H}_t)$. 
Typical acquisition functions include the expected improvement (EI)~\cite{marchuk1975,jones1998}, the upper confidence bound (UCB)~\cite{peter2003,srinivas2010,srinivas2012}, the entropy search~\cite{henniq2012,henrandez2014,ziw2017},  and the knowledge gradient~\cite{frazier2008,scott2011,wu2016}.

\begin{figure}
    \centering
    \includegraphics[width=0.82\linewidth,height=!]{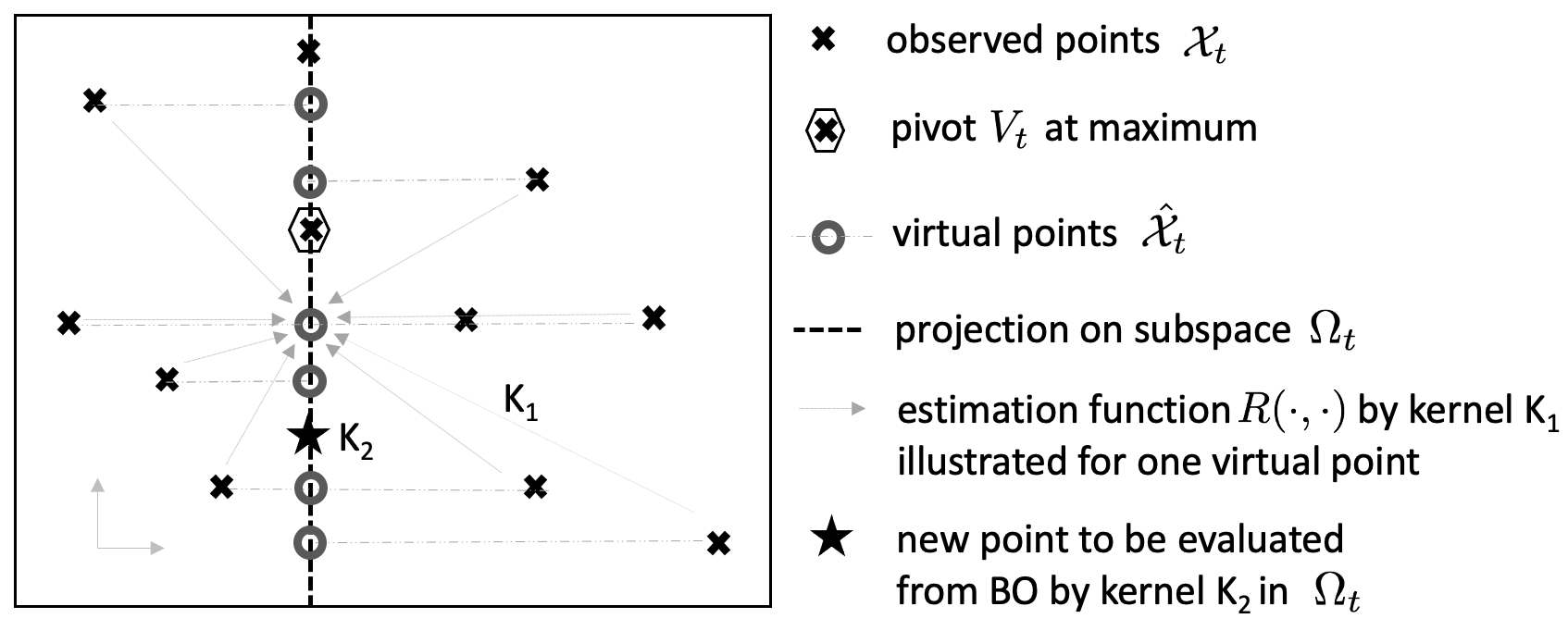}
	\caption{Illustrate of two-stage kernels; stage 1: subspace projection and function value estimation for virtual points using kernel $K_1$; stage 2: BO in $\Omega_t$ using kernel $K_2$}
	\label{fig:projection}
	\vspace{-3mm}
\end{figure}

Instead of directly computing the posterior distribution $p( \hat{f} | \mathcal{H}_t)$, 
%we use $p( \hat{f} | \hat{\mathcal{H}}_t)$, 
we replace the conditional events $\mathcal{H}_t$ by
 \begin{align*}
% 	 \vspace{-1.5mm}
 \hat{\mathcal{H}}_t = R\left( P_{\Omega_t} \left(\mathcal{X}_t\right), \mathcal{H}_t\right)=\left\{ \left( \hat{x}_i, \hat{y}_i  \right)\right\}_{i=1}^{t}
% p( \hat{f} | \mathcal{H}_t)  \coloneqq p\left[ \hat{f}_{\Omega_t}(x) | R\left( P_{\Omega_t} \left(\mathcal{X}_t\right), \mathcal{H}_t\right), x \in {\Omega}_t \right]
 \end{align*}
  with %interpolation $R(\cdot, \cdot)$ and
  %by the radial basis function (RBF) 
  %detailed in Eq.~\ref{eq:rbf}, 
  a projection function $P_{\Omega_t}(\cdot)$
%   \vspace{-3mm}
 \begin{equation}
    \label{eq:projection}
    P_{\Omega_t}(x)^{(j)}=
    \begin{cases}
			x^{(j)} & \text{if } j \in C_t \\
            V_t^{(j)} & \text{if } j \notin C_t
	 \end{cases} 
%   \vspace{-1mm}
 \end{equation}
 at coordinate $j$. It simply keeps the values of $x$ whose corresponding coordinates are in $C_t$ and replaces the rest by the corresponding values of $V_t$, as illustrated in Fig.~\ref{fig:projection}.
 % The subspace $\Omega_t$ is on a subset of randomly selected coordinates $C_t$.
 
%   \begin{figure}[htb]\vspace{2.5mm}
% 	\centering
% % 	\includegraphics[width=0.65\linewidth]{projection.png}
% 	\includegraphics[width=\columnwidth,height=!]{projection.png}
% 	\vspace{-6mm}
% 	\caption{Subspace projection and function value interpolation}\vspace{-6mm}
% 	\label{fig:projection}
% \end{figure}

Applying $P_{\Omega_t}(\cdot)$ on $\mathcal{X}_t$ and discarding duplicates generate a new set of distinct virtual points $\hat{\mathcal{X}}_t = \{\hat{x}_1,  \hat{x}_2,  \hat{x}_3, \cdots, \hat{x}_{\hat{t}}\}$, $\hat{x}_i \in \Omega_t \,\forall\,  1\leq i \leq \hat{t} \leq t$.
%\niv{What guarantees those virtual points to be unique ?}
In our implementation, the function values at $\hat{x}_i\in\hat{\mathcal{X}}_t$ are interpolated as $\hat{y}_i = R(\hat{x}_i, \mathcal{H}_t)$ using the standard radial basis function~\cite{buhmann_2003} due to its generality and existence of efficient implementations~\cite{Gumerov07}.  
Specifically, using a `multiquadric' kernel with length scales approximating the average distance between points, CobBO can efficiently fit the model in the full space. %Other efficient methods also exist; see Section~\ref{}.
%and the observed points in $\mathcal{H}_t$. 

\noindent \textbf{Alternative approach:}
note that a computation-efficient method in the first stage does not prevent us from using a sophisticated kernel, e.g., ARD Mat\'ern kernel, which however needs to be carefully used to reduce the computation time in the full space. For example,  one can keep using the same kernel across multiple iterations by remembering the updated parameters and in the meanwhile allowing only a small number of training steps in each iteration. This is possible since the first stage is conducted on the fixed and full space, where the same parameters and the kernel can be utilized and kept in memory. On the contrary, the second stage is on varying coordinate subspaces, where the parameters of the same kernel cannot be applied unanimously on different subspaces.  
%other kernels could also be applied in the full space. For example, a sophisticated kernel, ,  that remembers its parameters across different iterations can also be utilized, which however needs to limit the number of training steps in each iteration. 

%Interestingly,  many of the virtual points $\hat{\mathcal{X}}_t$ do not coincide with the already queried points $\mathcal{X}_t$.
%\niv{Evidence in later sections (ref) ? Citation ?}
% Together, the two-phase kernels are formed in the subspace $\Omega_t$, with the first step being the random projection in conjunction with the smooth interpolation and the second step being the Gaussian process regression with the chosen kernel, e.g., Automatic Relevance Determination (ARD) Mat\'{e}rn 5/2 kernel~\cite{matern}. 
%
% \begin{figure}[htb]
% \centering
% % 	\includegraphics[width=0.65\linewidth]{projection.png}
% 	\includegraphics[width=0.7\columnwidth,height=!]{figures/execution_time.png}
% 	\caption{Compare execution times of vanilla BO and CobBO}
% 	\label{fig:execution_time}
% \end{figure}
%
%It not only significantly reduces the GP regression time due to the efficiency of RBF~\cite{buhmann_2003} and the acquisition function optimization in low dimensions~\cite{josip2013}, but also eventually improves the model accuracy using the more sophisticated kernel applied on~${\Omega}_t$. 
%due to better estimations of the length scales~\cite{lengthscale} of the kernel functions. 

\subsection{Block coordinate ascent for subspace selection}\label{ss:block}%\vspace{-0mm}
% \vspace{-1.5mm}
%\textbf{Block coordinate ascent and subspace selection:}
%For Bayesian optimization, consider an infeasible assumption that each iteration can exactly maximize the function $f(x)$ in $\Omega_t$. This is not possible for one iteration but only if one can consistently query in $\Omega_t$, since the points converge to the maximum, e.g., under the expected improvement acquisition function with fixed priors~\cite{vazquez2010} and the convergence rate can be characterized~\cite{bull2011}. However, even with this infeasible assumption, it is known that coordinate ascent with fixed blocks can cause stagnation at a non-critical point~\cite{warga63,powell1972}.
 %, e.g.,  for non-differentiable~\cite{warga63} or non-convex functions~\cite{powell1972}. This motivates us to select a subspace with a variable-size coordinate block $C_t$ for each query.  
%A good coordinate block can help the iterations to escape the trapped non-critical points.  For example, one condition can be based on the result in~\cite{grippo00} that assumes $f(x)$ to be differentiable and strictly quasi-convex over a collection of blocks.  In practice, we do not restrict ourselves to these assumptions.  %expect more general conditions to apply.   % %

 We induce a preference distribution $\pi_t$ over the coordinate set $I$, and sample a variable-size coordinate block $C_t$ accordingly.  
This distribution is updated at iteration $t$ through a multiplicative weights update method~\cite{sanjeev12}. 
%\niv{The MW algorithm samples a single coordinate from $I$ according to $\pi_t$, such that $\sum_{i\in I}\pi_{t,i} =1$. CobBO samples a set of coordinates. How does it do so ? Is it that $\sum_{i\in I}\pi_{t,i} \neq 1$ and for every coordinate $i \in I$ include it in $C_t$ with probability $\pi_{t,i}$ ? If so, how exactly is $\pi_{t,i}$ normalized to induce a probability measure over the coordinate $i$ ?}
% based on which a coordinate block is sampled out. 
Specifically, 
% depending on whether a query in $\Omega_t$ improves $M_{t-1}$, i.e., the maximum of $f(x)$ on $\mathcal{X}_{t-1}$,  at iteration $t$ or not, 
the values of $\pi_t$ at coordinates in $C_t$ starts off uniform and increase in face of an improvement or decrease otherwise according to different multiplicative ratios $\alpha>1$ and $\beta>1$, respectively, 
%Option 1:
 \begin{align}
    \label{eq:multiplicative_update}
    w_{t, j} &= w_{t-1, j} \cdot
    \begin{cases}
			\alpha & \text{if } j \in C_t \text{ and } y_t > M_{t-1} \\
			1/\beta & \text{if } j \in C_t \text{ and } y_t \leq M_{t-1} \\
            1 & \text{if } j \notin C_t
	\end{cases} 
% 	\\
%     w_{0, j} &= \frac{1}{D} 
% 	\quad; \quad
%     \pi_{t,j} = \frac{w_{t,j}}{\sum_{j=1}^D w_{t,j}}  
 \end{align}
% Option 2: Algorithm~\ref{alg:mw}
%\input{algorithm/mw_algo}
with $w_{0, j}=\sfrac{1}{D}$ and $\pi_{t,j} = \sfrac{w_{t,j}}{\sum_{j=1}^D w_{t,j}}$. This update characterizes how likely a coordinate block can generate a promising search subspace. 
The multiplicative ratio $\alpha$ is chosen to be relatively large, e.g., $\alpha=2.0$, and $\beta$ relatively small, e.g., $\beta=1.1$, since the queries that improve the best observations $y_t > M_{t-1}$ happen more rarely than the opposite $y_t \leq M_{t-1}$.
%and $\beta=1.2$, so that the most recent queries can impact the next chosen coordinate set $C_{t+1}$ more influentially. 

%How to dynamically select the size $|C_t|$? It is known that Bayesian optimization works well for low dimensions~\cite{frazier2018}. Thus, we specify an upper bound for the dimension of the subspace (e.g. $|C_t|\leq 30$).
%  except when the number of queried points contained in a localized trust region $\tilde{\Omega}_t$ (see section~\ref{ss:auxiliary}) is smaller than a threshold (e.g. $200$), as for this small number of points the modest computational cost associated with the Gaussian process regression allows computations in higher dimensionality.
For selecting the size $|C_t|$, we specify an upper bound, e.g. $|C_t|\leq 30$, where $|C_t|$ can be any random number in a finite set $\mathcal{C}$. %of possible block sizes 
%  Empirically, we use a subset, 
%  e.g., $|C_t| \in \{2, 3, 5, 6, 9, 11, 13, 16, 19, 24, 27, 30, 35\}$.   
% This is different from the
Most existing methods partition 
 the coordinates into fixed blocks and select one according to, e.g., cyclic order~\cite{stephen2015}, random sampling or Gauss-Southwell~\cite{nutini2015}. 
\subsection{Backoff stopping rule for consistent queries}\label{ss:backoff}%\vspace{-1mm}
%\textbf{Backoff stopping rule for consistent queries:}
% Applying BO on $\Omega_t$ requires a 
% strategy to determine the number of consecutive queries for making a sufficient progress. 
% This strategy is based on previous observations, thus forming a stopping rule.  
% In principle, there are two different scenarios, exemplifying 
% exploration and exploitation, respectively.  
% Persistently querying a given subspace refrains from opportunistically exploring other coordinate combinations. 
% Abruptly shifting to different subspaces does not fully exploit the potential of a given subspace.  
Observe that only a fraction of the points in  $\hat{\mathcal{X}}_t \cap \mathcal{X}_t$ directly observe the exact function values. The function values on the rest ones in $\hat{\mathcal{X}}_t \backslash \mathcal{X}_t$ are estimated.
For the trade-off between the inaccurate estimations and the exact observations in $\Omega_t$, we design a stopping rule that determines the number of consistent queries in $\Omega_t$. The more consistent queries conducted in a given subspace, the more accurate observations could be obtained, albeit at the expense of a smaller remaining budget for exploring other regions.
%CobBO designs a heuristic stopping rule to address the above two scenarios.
% It takes the following into consideration: 1) a maximal query budget in each subspace grows with the total query budget and dimension; 2) a sufficient progress needs to be made in the subspace to avoid harvesting of marginal improvements due to local fluctuations. 
%The details are presented in Appendix~\ref{ss:stop}.

%It considers not only the number of consecutive queries that fail to improve the objective function but also other factors including the improved difference $M_t-M_{t-1}$, the point distance $||x_t - x_{t-1}||$, the query budget $T$ and the problem dimension $D$. 

For each iteration $t$, denote the relative improvement at iteration $t$ by \begin{align}
\Delta_t = \frac{y_t - M_{t-1}}{\max(\left|M_{t-1}\right|, 0.1)}. \nonumber
\end{align}
When looking backward in time from iteration $t$, we denote by $P_t$ the number of consecutive improvements ($\Delta_s>0, s\leq t$) and by $N_t$ the total number of consecutive queries in the same subspace as in $\Omega_t$, respectively.
We set
\begin{align*}
    C_{t+1} &= 
    \begin{cases}
            \text{a new block}, & N_t \geq \tau \text{ and } \Delta_t \leq 0.1  \text{ and } P_t \leq \xi \\
            C_t, & N_t < \tau \text{ or } \Delta_t > 0.1 \text{ or } P_t > \xi
    \end{cases}
\end{align*}
%$\tau$ represents the minimum number of consecutive queries in each subspace and $\xi$ is a threshold for $P_t$.
% \begin{align*}
%     % \tau = 
%     % \begin{cases}
%     %         1 & T\leq 100 \text{ and } D\leq 20 \\
%     %         5 & T\geq 5000 \text{ and } D\geq 50
%     % \end{cases}
%     % \qquad ; \qquad 
%     \xi = 
%     \begin{cases}
%             4 & \Delta_t < 0.05 \\
%             2 & 0.05 \leq \Delta_t \leq 0.1\\
%             0 & \Delta_t > 0.1
%     \end{cases}
% \end{align*}
where the value $\tau$ depends on the query budget $T$ and the problem dimension $D$, e.g.,
\begin{align*}
\tau = \frac{T}{1000} + \begin{cases}
            1 & D < 20 \\
            2 & 20 \leq D < 70 \\
            3 & 70 \leq D < 100 \\
            4 & 100 \leq D < 200 \\
            5 & 200 \leq D
    \end{cases} 
   \qquad %; \qquad 
        \xi = 
    \begin{cases}
            4 & \Delta_t < 0.05 \\
            2 & 0.05 \leq \Delta_t \leq 0.1\\
            0 & \Delta_t > 0.1
    \end{cases}
\end{align*}

%Effectively, 
This heuristic stopping rule is designed to take into account several considerations:
%listed in this section by:
%It takes the following into consideration: 1) a maximal query budget in each subspace grows with the total query budget and dimension; 2) a sufficient progress needs to be made in the subspace to avoid harvesting of marginal improvements due to local fluctuations.
\begin{enumerate}
    \item A maximal query budget ($\tau$) in each subspace grows with the total query budget ($T$) and dimension ($D$).
    \item  A sufficient progress ($\Delta_t$) needs to be made in the subspace to avoid only harvesting marginal improvements due to local fluctuation. The more significant progress the more consecutive improvements ($\xi$) are allowed in this subspace.
\end{enumerate}

This heuristic stopping rule is robust to all the problems presented in this work and to many other that we have tested.

\subsection{Theoretical motivation for the subspace selection policy} \label{ss:analysis}
This section aims at providing some theoretical guidance for specifying the proposed block coordinate selection scheme in section~\ref{ss:block} rather than it being merely a heuristic.
It can be viewed as a combinatorial mixture of experts problem~\cite{cesa2006prediction}, where each coordinate is a single expert and the forecaster aims at choosing the best combination of experts in each step. Under this view, we bound the regret of our selection method on an intuitive surrogate loss function with respect to the policy of selecting the best (unknown) block of coordinates at each step.
This is complementary to the regret analysis of the optimization procedure performed at each subspace. Specifically, the conventional regret analysis associated with BO, with respect to the value of the objective function, is applicable for each specific subspace, accounting for the projection of points to this subspace. Here we focus on justifying the coordinate selection part alone.

Following the standard framework, we compare with a fixed optimal choice $\mathcal{I}^*$ for the block of coordinates to pick at all steps. This block is characterized by improving the objective function for the largest number of times among all the possible coordinate blocks when performing Bayesian optimization. 
%We design the following loss function. 
For any coordinate subset $\mathcal{A}$,
%and a sequence of threshold values $\{\lambda_t\}_{t=0}^{T-1}$ given in advance by some oracle, 
we define the following loss function at time $t$, for coordinate $i$,
% \begin{align} \label{eq:alpha_beta_loss}
%     \ell_{t,i}(C_t, y_t) =
%     \begin{cases}
% 		-
%         % \frac{1}{\eta}
% 		\log(\tilde{\alpha}) & \text{if } i \in C_t \text{ and } y_t > M_{t-1} \\
% % 		\frac{1}{\eta}
% 		\log(\tilde{\beta}) & \text{if } i \in C_t \text{ and } y_t \leq M_{t-1} \\
%         0 & \text{if } i \notin C_t
%     \end{cases} 
%     \quad ; \quad 
%     \tilde{\alpha},\tilde{\beta}>1
% \end{align}
\begin{align} \label{eq:alpha_beta_loss}
    \ell_{t,i}(\mathcal{A}) =
    \begin{cases}
		-
        % \frac{1}{\eta}
		\log(\tilde{\alpha}) & \text{if } i \in \mathcal{A} \text{ and } y_t > M_{t-1} \\
% 		\frac{1}{\eta}
		\log(\tilde{\beta}) & \text{if } i \in \mathcal{A} \text{ and } y_t \leq \lambda_t \\
        0 & \text{if } i \notin \mathcal{A}
    \end{cases} 
\end{align}
with $\tilde{\alpha},\tilde{\beta}>1$, where both $y_t$ and $M_{t-1}$ are fully determined by the previously selected coordinate subset $C_1, C_2, \cdots, C_{t-1}, C_t$.  
All the coordinates participating in the selected block incur the same loss that effectively rewards these coordinates for improving the objective and penalizes these for failing to improve the objective. All other coordinates that are not selected receive a zero loss and remain untouched.

Note that $\tilde{\alpha}$ and $\tilde{\beta}$ express the extent of reward and penalty, e.g. for $\tilde{\alpha}=\tilde{\beta}=e$ we have %balanced 
losses of $\ell_{t,i} \in\{-1, 1, 0\}$. %respectively.
% \begin{align} \label{eq:alpha_beta_loss_ones}
%     \ell_{t,i} =
%     \begin{cases}
% 		-
%         % \frac{1}{\eta}
% 		1 & \text{if } i \in C_t \text{ and } y_t > M_{t-1} \\
% % 		\frac{1}{\eta}
% 		1 & \text{if } i \in C_t \text{ and } y_t \leq M_{t-1} \\
%         0 & \text{if } i \notin C_t
%     \end{cases}
% \end{align}
Yet, $\tilde{\alpha}$ is chosen to be larger than $\tilde{\beta}$, since the frequency of improving the objective is expected to be smaller.%since improving the objective can be rather hard and thus the queries that improve the best observations $y_t > M_{t-1}$ happen more rarely than the opposite $y_t \leq M_{t-1}$. 

The loss received by the forecaster is to reflect the same motivation. This is done by averaging the losses of the individual coordinates in the selected block, so that the size of the block does not matter explicitly, i.e. a bigger block should not incur more loss just due to its size but only due to its performance. Such that for each coordinate block $\mathcal{I}_t
\subset\mathcal{I}=\{1,\cdots,D\}$ selected at time step $t$, the loss incurred by the forecaster is
 $L_{t,\mathcal{I}_t} =\frac{1}{|\mathcal{I}_t|} \sum_{i\in\mathcal{I}_t}\ell_{t,i}$.
% \begin{equation}
%     L_{t,\mathcal{I}_t} =\frac{1}{|\mathcal{I}_t|} \sum_{i\in\mathcal{I}_t}\ell_{t,i} = \bar{\ell}_{t,\mathcal{I}_t}
% \end{equation}
This is also the common loss incurred by all the coordinates participating in that block.

In each step we have the following multiplicative update rule of the weights associated with each coordinate
 \begin{align}
    \label{eq:multiplicative_weight_update}
    w_{t, i}
	&= 
	w_{t-1, i} \cdot e^{-\eta\ell_{t,i}(C_t; y_t, M_{t-1})}
	=
    w_{t-1, i} \cdot
    \begin{cases}
			\tilde{\alpha}^\eta & \text{if } i \in C_t \text{ and } y_t > M_{t-1} \\
			1/\tilde{\beta}^\eta & \text{if } i \in C_t \text{ and } y_t \leq M_{t-1} \\
            1 & \text{if } i \notin C_t,
	 \end{cases} 
% 	 =
%     w_{t-1, i} \cdot
%     \begin{cases}
% 			\alpha & \text{if } i \in C_t \text{ and } y_t > M_{t-1} \\
% 			1/\beta & \text{if } i \in C_t \text{ and } y_t \leq M_{t-1} \\
%             1 & \text{if } i \notin C_t
% 	 \end{cases}
 \end{align}
which, by setting $\alpha=\tilde{\alpha}^\eta$ and $\beta=\tilde{\beta}^\eta$, yields the update rule in Eq.~(\ref{eq:multiplicative_update}).
% Denote all possible block sizes by $\mathcal{C}$ and the set of all possible coordinate blocks of size $c\in\mathcal{C}$ by $\mathcal{S}_c$ and its size $|\mathcal{S}_c|=\frac{D!}{c!(D-c)!}={D \choose c}$. Denote by $p_c$ the probability of choosing a certain block size $c\in\mathcal{C}$, such that $p_c\geq 0$ and $\sum_{c\in\mathcal{C}}p_c=1$. 

The probability $\tilde{\pi}_{t,\mathcal{I}_t}$ of selecting a certain coordinate block $\mathcal{I}_t$ is induced by $\pi_t$ as specified next. 
%Assume the threshold values $\{\lambda_t\}_{t=0}^{T-1}$ are given in advance by some oracle and at each time step some oracle $f_t(\cdot)$ yields $f_t(\mathcal{I}_t)=y_t$. %without replacement (explicitly put in Appendix~\ref{sec:regret_analysis}).
Thus the expected cumulative loss of the forecaster is:
$$L_T= \sum_{t=1}^T\sum_{c\in\mathcal{C}}\sum_{\mathcal{I}_t\in \mathcal{S}_c}\tilde{\pi}_{t,\mathcal{I}_t}\cdot\frac{1}{|\mathcal{I}_t|} \sum_{i\in\mathcal{I}_t}\ell_{t,i}$$%(\mathcal{I}_t;f_t(\mathcal{I}_t), \lambda_t)$$

Assume that the best coordinate block is $\mathcal{I}^*$, then the corresponding cumulative loss is:
\begin{align*}
%L_T^* &= \sum_{t=1}^T L_{t,\mathcal{I}^*} \nonumber\\
L_T^* &= \sum_{t=1}^T L_{t,\mathcal{I}^*}=\sum_{t=1}^T\frac{1}{|\mathcal{I}^*|} \sum_{i\in \mathcal{I}^*}\ell_{t,i}%(\mathcal{I}^*;f_t(\mathcal{I}^*), \max_{0\leq \tau<t}f_{\tau}(\mathcal{I}^*)). \nonumber
% = \sum_{t=1}^T\bar{\ell}_{t,\mathcal{I}^*} \nonumber
\end{align*}

We hence aim at bounding the regret $\mathcal{R}_T= L_T-L_T^*$. 
%For this purpose we bound the regret with respect to any arbitrary sequence of selected coordinate blocks.

% \textbf{Theorem 1} 
\begin{theorem} 
\label{theo:regret_comb}
Sample from the combinatorial space of all possible coordinate blocks $\mathcal{I}_t \in \bigcup_{c\in\mathcal{C}}\mathcal{S}_c$ with probability 
% $\tilde{\pi}_{t,\mathcal{I}_t} = \sfrac{\prod_{i\in \mathcal{I}_t}
% w_{t,i}^{\sfrac{1}{|\mathcal{I}_t|}}
% % \sqrt[|\mathcal{I}_t|]{w_{t,i}}
%     }
%     {
%   \sum_{c\in\mathcal{C}}\sum_{\mathcal{I}_t\in\mathcal{S}_c}\prod_{j\in \mathcal{I}_t}
%   w_{t,j}^{\sfrac{1}{|\mathcal{I}_t|}}
%     % \sqrt[|\mathcal{I}_t|]{w_{t,j}}
%     }
% $. 
$\tilde{\pi}^{c}_{t,\mathcal{I}_t} =\sfrac{ 
\prod_{i\in \mathcal{I}_t}
\tilde{w}_{t,\mathcal{I}_t}
}{
   \sum_{c\in\mathcal{C}}\sum_{\hat{\mathcal{I}}\in\mathcal{S}_c}\prod_{j\in \hat{\mathcal{I}}}
   \tilde{w}_{t,\hat{\mathcal{I}}}
   }
$. Then the update rule in Eq.~(\ref{eq:multiplicative_update}) with $\alpha=\tilde{\alpha}^\eta$, $\beta=\tilde{\beta}^\eta$ and
$\eta=\log(\tilde{\alpha}\tilde{\beta})^{-1}\sqrt{T^{-1}|\mathcal{C}|D\log(D)}$ yields
 \begin{equation}\label{eq:theorem_1}
     \mathcal{R}_T
     \leq 
      \mathcal{O}\left(\log(\tilde{\alpha}\tilde{\beta})\cdot \sqrt{T|\mathcal{C}|D\log(D)}\right),
 \end{equation}
 where 
%  $\tilde{w}_{t,\mathcal{I}_t}=\sqrt[|\mathcal{I}_t|]{\prod_{i\in \mathcal{I}_t}w_{t,i}}$ 
 $\tilde{w}_{t,\mathcal{I}_t}=\prod_{i\in \mathcal{I}_t}w_{t,i}^{\sfrac{1}{|\mathcal{I}_t|}}$ 
 is the geometric mean of the weights for block $\mathcal{I}_t$.
\end{theorem}
% \vspace{-2mm}
 The upper bound in Eq.~(\ref{eq:theorem_1}) is tight, as the lower bound can be shown to be of $\Omega(\sqrt{T\log(N)})$ \cite{haussler1995tight} where the number of experts is $N=\sum_{c\in\mathcal{C}}\mathcal{S}_c\leq D^{|\mathcal{C}|D}$ in our combinatorial setup, as typically $|\mathcal{C}|\ll D$.
 
 In practice, the direct sampling policy introduced in Theorem~\ref{theo:regret_comb} involves high computational costs due to the exponential growth of combinations in $D$. Thus CobBO suggests an alternative computationally efficient sampling policy with a linear growth in $D$.
% \textbf{Theorem 2} 
\begin{theorem} \label{theo:regret_without_replacement} Sample a block size $c\in\mathcal{C}$ with probability $p_c$ and $c$ coordinates without replacement according to $\pi_t$. Assume $\mathcal{C}\supset\{1\}$, then 
the update rule in Eq.~(\ref{eq:multiplicative_update}), with $\alpha=\tilde{\alpha}^\eta$, $\beta=\tilde{\beta}^\eta$ and\\
$\eta=\sqrt{\frac{\log(D)}{T(\log(\tilde{\alpha}\tilde{\beta})^2 -\log(p_1))}}\geq 1$ yields
 \begin{equation}\label{eq:regret_without_replacement}
     \mathcal{R}_T
     \leq 
      \mathcal{O}\left(\sqrt{(\log(\tilde{\alpha}\tilde{\beta})^2 -\log(p_1))} \cdot \sqrt{T\log(D)})\right),
 \end{equation}
 where $p_c>0$ for all $c\in\mathcal{C}$ and $\sum_{c\in\mathcal{C}}p_c=1$.
\end{theorem}
% \vspace{-2mm}
%, e.g., uniformly set $p_c\equiv|\mathcal{C}|^{-1}$.%\sfrac{1}{|\mathcal{C}|}$.
The proof and detailed sampling policy are in appendix~\ref{sec:regret_analysis}. The regret upper bound in Eq.~\ref{eq:regret_without_replacement} is tight, as the lower bound for an easier setup can be shown to be of $\Omega(\sqrt{T\log(D)})$ \cite{haussler1995tight}. 
 The implication on $\eta$ is valid only for settings of a high dimension and low query budget. In particular, CobBO is designed 
 for this kind of problems. 
% Moreover, interestingly, although the effective number of combinations of coordinates is $|\mathcal{S}|\leq  |\mathcal{C}|\cdot(D!)$, the regret upper bound in \ref{eq:regret} shows to grow with $\mathcal{O}(\sqrt{\log(D)})$ rather than $\mathcal{O}(\sqrt{\log(|\mathcal{C}|)+\log(D!)})\sim\mathcal{O}(\sqrt{D\log(D)})$ for large $D>>|\mathcal{C}|$ due to the Stirling's approximation~\cite{pearson1924historical}. This is due to adapting the preference probability $\pi_{t,i}$ for each coordinate rather than the one for each possible combination of coordinates $\hat{\pi}_{t,\mathcal{I}_t}$, as the later is derived from the former.
%\textbf{Remark:} 
Similar analysis and results follow when incorporating consistent queries from section~\ref{ss:backoff} and sampling a new coordinate block once every several steps. This is done by effectively performing less steps of aggregated temporal losses, as shown in appendix~\ref{sec:regret_analysis_consistent_queries}.

\section{Numerical Experiments}\label{sec:num_exp}

% \begin{figure*}[!ht]%\vspace{-2mm}
% \begin{center}
%   \includegraphics[width=0.95\linewidth,height=!]{figures/ablation.png}\vspace{-5mm}
% %   \includegraphics{lunar-robot.png}
% \end{center}
%   \caption{Ablation study using Rastrigin on $[-5,10]^{50}$ with $20$ initial samples}
%   \label{fig:ablation}
% \end{figure*}
This section presents detailed ablation studies of the key components %presented in Section~\ref{sec:algorithm} 
and comparisons with other algorithms.
The specifications of the testbed are as follows: Intel(R) Xeon(R) CPU E5-2682 v4 2.50GHz, Memory 32GB, GPU NVIDIA Tesla P100 PCIe 16GB.

% \vspace{-2mm}
\subsection{Ablation study and empirical analysis}
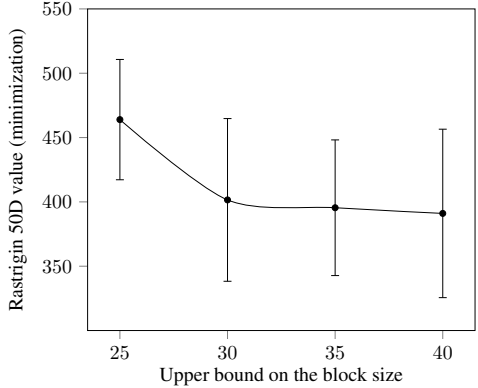
\begin{figure}[h!]
\centering
\begin{adjustbox}{width=0.45\textwidth}
\begin{tikzpicture}
\begin{axis}[
  ymin=300, ymax=550, 
  ytick={350,400,...,550}, 
  ytick align=outside, ytick pos=left,
  xtick={25,30,35,40}, 
  xtick align=outside, 
  xtick pos=left,
  xlabel=Upper bound on the block size,
  ylabel={Rastrigin 50D value (minimization)},
  legend pos=north west,
  legend style={draw=none}]
\addplot+[
  black, mark options={black, scale=0.75},
  smooth, 
  error bars/.cd, 
    y fixed,
    y dir=both, 
    y explicit
] table [x=x, y=y,y error=error, col sep=comma] {
    x,  y,       error
    25,  463.96,  46.76
    30,  401.53,  63.24
    35,  395.45,  52.73
    40,  391.04,  65.47
};
\end{axis}
\end{tikzpicture}
\end{adjustbox}
\caption{Impact of the block size upper bound on the best function values for Rastrigin on $[-3, 4]^{50}$}
\label{fig:upper_bound}
\end{figure}

\begin{figure*}[!h]%\vspace{-2mm}
\begin{center}
  \includegraphics[width=1\linewidth,height=!]{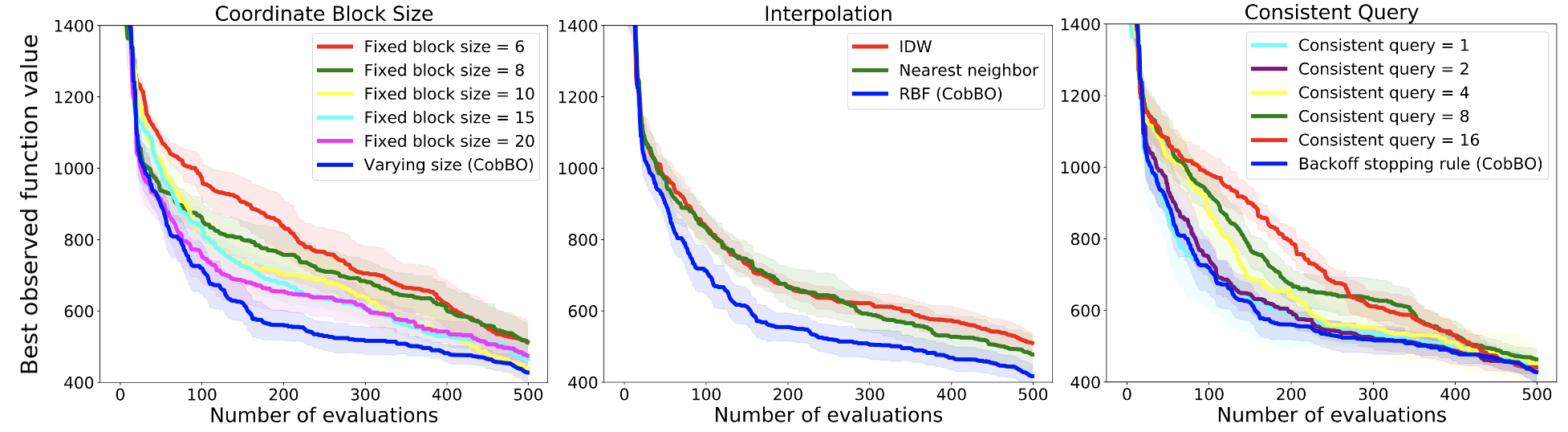}%\vspace{-5mm}
  \vspace{-1mm}
\end{center}%\vspace{1mm}
  \caption{Ablation study using Rastrigin on $[-5,10]^{50}$ with $20$ initial random samples}
  \vspace{-1mm}
  \label{fig:ablation_trio}
\end{figure*}

\begin{figure}
    \centering
  \includegraphics[width=0.5\textwidth,height=!]{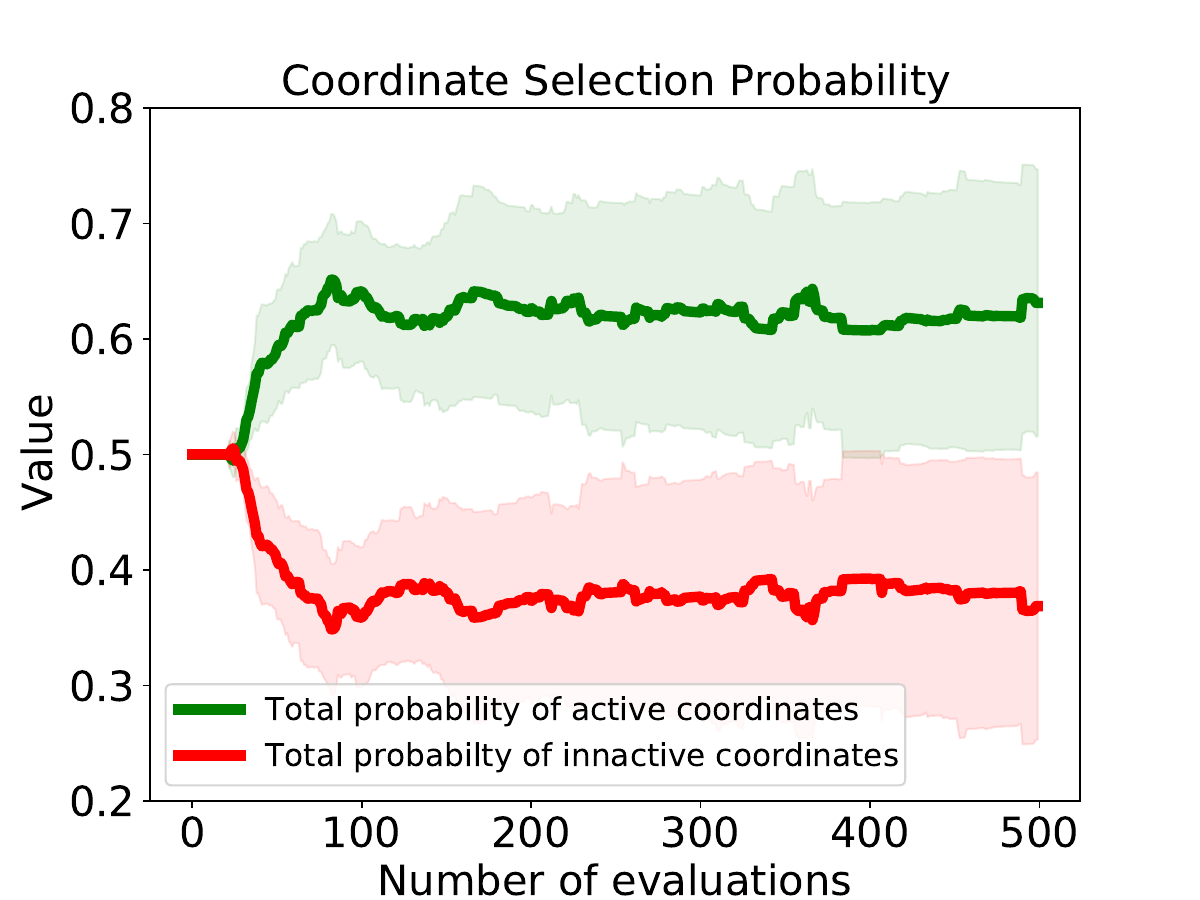}
  \vspace{-2mm}
  \caption{The preference probability focuses on active coordinates}% as the entropy decreases}
  \label{fig:select_prob}
  \vspace{-2mm}
\end{figure}

% \vspace{-1.5mm}
Ablation studies are designed to study the contributions of the key components in 
Algorithm~\ref{alg:top} by experimenting with the Rastrigin function on $[-5,10]^{50}$ with 20 initial points. %The best performing run out of 5
Confidence intervals ($95\%$) over $10$ independent experiments for each 
configuration are presented in Fig.~\ref{fig:upper_bound}, ~\ref{fig:ablation_trio}~and ~\ref{fig:select_prob}. 
%for $500$ queries.

\textbf{The upper bound of the block sizes: }\label{sec:upper_bound_ablaition}
At each iteration, the block size $|C_t|$ of CobBO is uniformly sampled from a set formed through capping the elements from $\{1,4,6,8,12,14,16,22,24,26,30\}$ by the dimension $D$ of the problem. Hence the average block size is about $15$, the lower bound is $1$ and the upper bound is $30$. This set is chosen to prefer relatively lower dimensions and works well for the problems we experimented with. 
In Fig.~\ref{fig:upper_bound} we present an ablation study focusing on the selection of the upper bound of this set, which plots the means and variances of the best searched function values for Rastrigin on $[-3, 4]^{50}$. 
Considering that 
the differences of the mean values of the best obtained minimization solutions are small compared to the standard deviations, we conclude that the algorithm is not very sensitive to the choice of the upper bound, while higher values are slightly favourable, as expected, yet require more computation.
%for $10$ different runs. 
%Specifically, for Rastrigin 50D, setting the upper bound to be 25, 30, 35 and 40 we get the mean best values of 463.96, 401.53, 395.45 and 391.04 respectively for 5 runs each with standard deviations of 46.76, 63.24, 52.73 and 65.47 respectively. 

\textbf{Coordinate blocks of a varying size: } 
%As a variant of coordinate ascent, 
CobBO selects a block of coordinates $C_t$ of a varying size, as described in Section~\ref{ss:block}. 
The above ablation study in Fig.~\ref{fig:upper_bound} shows that CobBO is quite robust to the upper bound of the block size $|C_t|$.
Fig.~\ref{fig:ablation_trio} (left) shows that a varying size is better than a fixed one. 
 Furthermore, although the average block size of CobBO is $15$ in this setting, it enjoys both the fast exploration of larger block sizes (e.g. $22$) and efficient exploitation of smaller block sizes (e.g. $6$).

% \vspace{-0.5mm}
\textbf{RBF interpolation in the first stage: }
%\textbf{RBF interpolation:} 
RBF calculation is time efficient, which is very beneficial in high dimensions. Fig.~\ref{fig:motivation} (left) shows the computation time of plain Bayesian optimization compared to CobBO's. While the former applies GP regression using the Mat\'{e}rn kernel in the high dimensional space directly, the later applies RBF interpolation in the high dimensional space and GP regression with the Mat\'{e}rn kernel in the low dimensional subspace. This two-stage kernel method leads to a significant speed-up. Other time efficient alternatives are, e.g., the inverse distance weighting~\cite{idw} and the simple approach of assigning the value of the observed nearest neighbour.  
Fig.~\ref{fig:ablation_trio} (middle) shows that RBF is more favorable.

% \vspace{-0.0mm}
\textbf{Backoff stopping rule: } CobBO applies a stopping rule to query a variable number of points in subspace $\Omega_t$ (Section~\ref{ss:backoff}).
%which balances between accurate observations and inaccurate estimations in $\Omega_t$. 
To validate its effectiveness, we compare it with schemes that use a fixed budget of queries for $\Omega_t$. Fig.~\ref{fig:ablation_trio} (right) shows that the stopping rule yields superior results. Specifically, it enjoys both fast exploration of small query budget in each subspace (e.g. $1$,$2$) and efficient exploitation of large ones (e.g. $16$). Note that for different problems the best fixed number of consistent queries vary but the backoff stopping rule can adaptively achieve a good performance. 
%better than others with fixed query budgets. 

% \vspace{-0.5mm}

% \vspace{-0.5mm}
% \textbf{Forming trust regions on two different time scales:}
% CobBO alternates between coarse and fine trust regions on slow and fast time scales, respectively (Section~\ref{ss:2tr}). 
% Figure~\ref{fig:ablation} (c) compares CobBO with two other schemes: without any trust regions and forming only coarse trust regions. Two time scales show better results. 

% \vspace{-0.5mm}
% \textbf{Escaping trapped optima:} 
% %CobBO applies two methods to escape trapped local optima.
% Figure~\ref{fig:ablation} (d) shows that the way CobBO escapes local optima (Section~\ref{ss:escape}) by decreasing $M_{t-1}$ and setting $V_{t}$ as a selected random point is beneficial.

% \begin{wrapfigure}{r}{0.5\textwidth}\vspace{-4mm}
% \begin{center}
%   \includegraphics[width=0.45\textwidth,height=!]{figures/select_prob.pdf}\vspace{-2mm}
% \end{center}
%   \caption{The preference probability focuses on active coordinates as the entropy decreases}
%   \label{fig:select_prob}
% \end{wrapfigure}

\textbf{Preference probability over coordinates: } For demonstrating the effectiveness of coordinate selection (Section~\ref{ss:block}), we artificially let the function value only depend on the first $25$ coordinates of its input and ignore the rest. It forms two separate sets of active and inactive coordinates, respectively. We expect CobBO to refrain from selecting inactive coordinates. Fig.~\ref{fig:select_prob} shows the overall preference probability $\pi_t$ for picking active ($\sum_{i=1}^{25}\pi_{t,i}$) and inactive coordinates ($\sum_{i=26}^{50}\pi_{t,i}$) at each iteration $t$. We see that the preference distribution concentrates on the active coordinates.

\subsection{Comparisons with other methods}\label{s:exp}
%After tuning the hyper-parameters of CobBO over a number of commonly used benchmarks, 
We fix a default configuration for CobBO identically across all of the experiments. 
%The values are specified in the supplementary materials together with more experiments. 
Extensive experiments show that CobBO performs on par or outperforms a collection of state-of-the-art methods. 
%This further demonstrates the robustness of CobBO.
Most of the experiments are conducted using the same settings as in TurBO~\cite{turbo2019}, where it is compared with a comprehensive list of baselines, including BFGS, BOCK~\cite{bock2018}, BOHAMIANN, CMA-ES~\cite{cmaes}, BOBYQA, EBO~\cite{wang18aistats}, GP-TS, HeSBO~\cite{chaudhuri2019}, Nelder-Mead and random search. 
To avoid repetitions, we only show TuRBO and CMA-ES that achieve the best performance among this list, and additionally compare with BADS~\cite{luigi2017},  
% HDBBO~\cite{zi2017}, SIR~\cite{miao2019},
Tree Parzen Estimator (TPE)~\cite{TPE2011} and Adaptive TPE (ATPE)~\cite{ATPE}. 
%The python code of the experiments will be made publicly available, together with the implementation of CobBO.  
Since CobBO is designed for high dimensional problems, we benchmark the performance in Section~\ref{ss:highD} in high dimensions. To show that it also works in low dimensions, we conduct the low dimensional tests in Section~\ref{ss:lowDtest}. 
As mentioned in Section~\ref{sec:related_work}, the embedding algorithms (e.g., REMBO~\cite{ziyuw2016} and ALEBO~\cite{letham2020}) and CobBO are based on different assumptions, which are
compared in Section~\ref{ss:alebo}. Section~\ref{ss:linebo} presents the comparison with LineBO~\cite{linebo}.
%and thus complement each other, 
%We repeat each experiment independently for 30 times to get the 95\% confidence intervals. 
%, and d-KG~\cite{wujian2017}.
%Though LineBO~\cite{linebo} and DROPOUT~\cite{dropoutbo} are also based on subspace selection,
%they do not show comparable performance.  
%Confidence intervals are computed with the results of 30 independent experiments.

\subsubsection{High dimensional tests}\label{ss:highD}

Since the duration of each experiment in this section is long, confidence intervals ($95\%$) over repeated 10 independent experiments for each problem are shown.

 %\vspace{0.25cm}
     \begin{figure}[!htb]
   \centering
      \includegraphics[width=0.8\columnwidth,height=!]{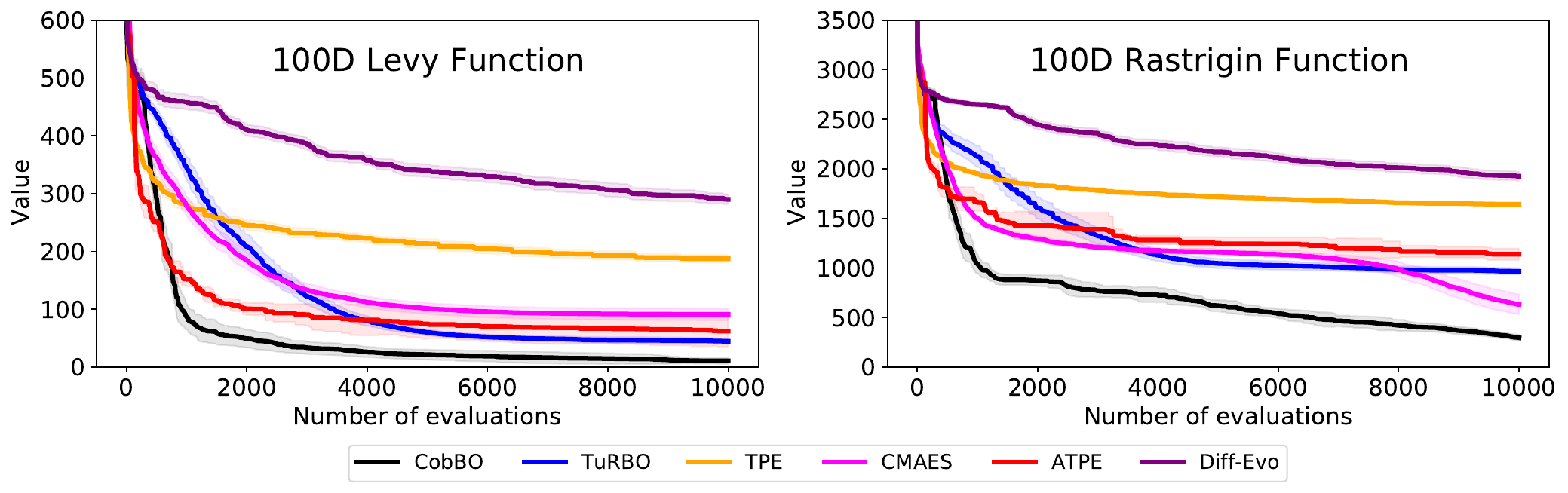}
   \vspace{-0.3cm}
   \caption{Performance over high dimensional synthetic problems: Levy (left) and Rastrigin (right)}
   \label{fig:100d} %\vspace{5mm}
 \end{figure}
\noindent \emph{The 100 dimensional synthetic black-box functions (minimization):} %\label{sec:100d}
We minimize the Levy and Rastrigin functions on $[-5, 10]^{100}$ with $300$ initial points.  These two problems are challenging since they have no redundant dimensions. % in high dimensions. 
TuRBO is configured with $1$ trust regions and a batch size of $100$.
 Fig.~\ref{fig:100d} (left) shows that CobBO can greatly reduce the trial complexity. 
  For Levy and Rastrigin, CobBO surpasses the final solutions of all the other methods within $2,000$ and $5,000$ trials for a total budget of $10,000$ trials, respectively. 
  REMBO is especially compared in Section~\ref{ss:alebo}. 
%  For Levy, it finds solutions close to the final one within $1,000$ trials, and eventually reach the best solution among all the algorithms tested.
%  For Rastrigin, within $1,000$ trials CobBO surpasses the final solutions of all the other methods.  REMBO is especially compared in Section~\ref{ss:alebo}. 
 %eventually with a large margin.
 %is excluded in this comparison, since it is designed for problems with low effective dimensions but Ackley 100D is not. See the comparisons in Section~\ref{ss:alebo}. 

In order to highlight the difference of the running time, we test Ackley 200D with $10,000$ trials. For a fair comparison, we change the configure so that both TurBO and CobBO have the same batch size of $1$. CobBO runs for $12.8$ CPU hours and TuRBO-1 runs for more than $80$ CPU hours or $9.6$ \emph{GPU} hours. Other methods either take too long to make progress or find far worse solutions.

 \vspace{0.35cm}
\noindent \emph{Additive latent structure (minimization):}
As mentioned in Section~\ref{sec:related_work}, additive latent structures have been explored for tackling challenges in high dimensions.
%which however incur a high computational cost~\cite{chaudhuri2019}.   %For $x=(x_1, x_2, x_3, x_4)$,  
We construct two additive functions. The first one has 36 dimensions, defined as  
 $f_{36}(x)=\rm{Ackley}(x_1) + \rm{Levy}(x_2) + \rm{Rastrigin}(x_3) + \rm{Hartmann}(x_4)$, where the first three terms express the exact functions and domains described in Section~\ref{ss:lowDtest},  with the Hartmann function defiend over $[0, 1]^{6}$. 
 The second has 56 dimensions, defined as 
 $f_{56}(x) = \rm{Ackley}(x_1) + \rm{Levy}(x_2) + \rm{Rastrigin}(x_3) + \rm{Hartmann}(x_4) +\rm{Rosenbrock}(x_5)+\rm{Schwefel}(x_6)$, 
 where the first four terms are the same as those of $f_{36}$, with the Rosenbrock and Schwefel functions defined over $[-5,10]^{10}$ and $[-500,500]^{10}$, respectively. 

\begin{figure*}[htb]\vspace{-3mm}
\begin{center}
  \includegraphics[width=1.0\linewidth,height=!]{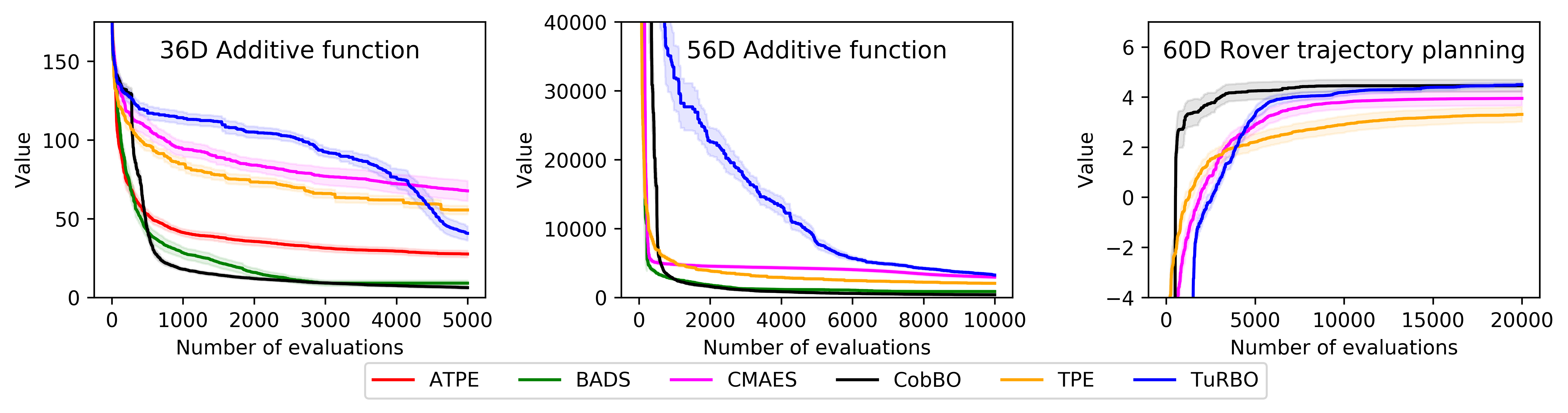}
\end{center}
  \caption{Performance over medium-size dimensional problems: 36D (left) and 56D (middle) additive functions and the 60D rover trajectory planning (right)}%\vspace{-1.5mm}
  \label{fig:highDims}
\end{figure*}

We compare CobBO with TPE, ATPE, BADS, CMA-ES and TuRBO, each with $100$ initial points. 
Specifically, TuRBO is configured with 15 trust regions and a batch size 50 for $f_{36}$ and $100$ for $f_{56}$. 
ATPE is excluded for $f_{56}$ as it takes more than 24 hours per run to finish. 
The results are shown in Fig.~\ref{fig:highDims}, where CobBO quickly finds the best solutions for both $f_{36}$  and $f_{56}$.

As shown in Fig.~\ref{fig:highDims}, CobBO finds the best solutions for both $f_{36}$  and $f_{56}$. 
BADS performs closely to CobBO. ATPE outperforms TPE, TuRBO and CMA-ES on $f_{36}$. 
TuRBO surpasses TPE and CMA-ES on $f_{36}$ eventually, while TPE and CMA-ES converge faster than TuRBO on $f_{56}$.

% \begin{figure}[htb]\vspace{-2mm}
% \begin{center}
%   \includegraphics[width=0.75\columnwidth,height=!]{figures/medium_v.png}\vspace{-3mm}
% %   \includegraphics{highDims.png}
% \end{center}
%   \caption{Performance over medium-size dimensional problems: 56D additive functions (upper) and the 60D rover trajectory planning (lower)}\vspace{-1.5mm}
%   \label{fig:highDims}
% \end{figure}

 %\vspace{0.25cm}
\noindent \emph{Rover trajectory planning (maximization):} 
This problem (60 dimensions) is introduced in~\cite{wang18aistats}. 
The objective is to find a collision-avoiding trajectory of a sequence consisting of 30 positions in a 2-D plane. 
%$[0,1]^{2}$. 
We compare CobBO with TuRBO, TPE and CMA-ES with a budget of $20,000$ evaluations and
$200$ initial points. 
TuRBO is configured with $15$ trust regions and a batch size of $100$, as in~\cite{turbo2019}. 
ATPE, BADS and REMBO are excluded for this problem and the following ones, as they all last for more than 24 hours per run. The result is shown in Fig.~\ref{fig:highDims}. CobBO reaches the best solution with fewer evaluations than TuRBO, while TPE and CMA-ES reach inferior solutions.

\subsubsection{Low dimensional tests}\label{ss:lowDtest}
 To evaluate the performance of CobBO on low dimensional problems, we use two more challenging problems of lunar landing~\cite{turbo2019}  and robot pushing~\cite{wang18aistats}, as well as classic synthetic black-box functions~\cite{TestProblems2013},  by following the setup in~\cite{turbo2019} for most of the experiments. Confidence intervals ($95\%$) over repeated 30 independent experiments for each problem are shown.

%  \vspace{0.25cm}
% \noindent \emph{The 30-dimensional classic functions:}
% We compare CobBO with TuRBO, BADS, TPE, ATPE and CMA-ES on the 30 dimensional versions of the Ackley, Levy and Rastrigin functions 
% introduced in Section \ref{ss:lowDtest}. %(except the Hartmann function that is defined to be fixed 6 dimensional)

% %   \begin{figure}[!ht]\vspace{-0mm}
% %   \centering
% %   \includegraphics[width=1.\columnwidth,height=!]{supplementary/30D-tests.png}\vspace{-3mm}
% %   \caption{Medium dimensional problems: Ackley (left), Levy (middle) and Rastrigin (right)} 
% %   \label{fig:30D-tests}
% %   \vspace{0.3cm}
% %  \end{figure}
% As shown in Fig.~\ref{fig:synthetic}, CobBO finds the global optima of Ackley the Levy, and the best results for Rastrigin. 
% BADS is competitive with CobBO on Ackley and Levy, while it performs next to CobBO on Rastrigin.  
% CMA-ES outperforms TuRBO, TPE and ATPE on Ackley, and is comparable to TPE on the other two problems. 

\begin{figure*}[htb]
\vspace{-2mm}
\begin{center}
  \includegraphics[width=0.8\linewidth,height=!]{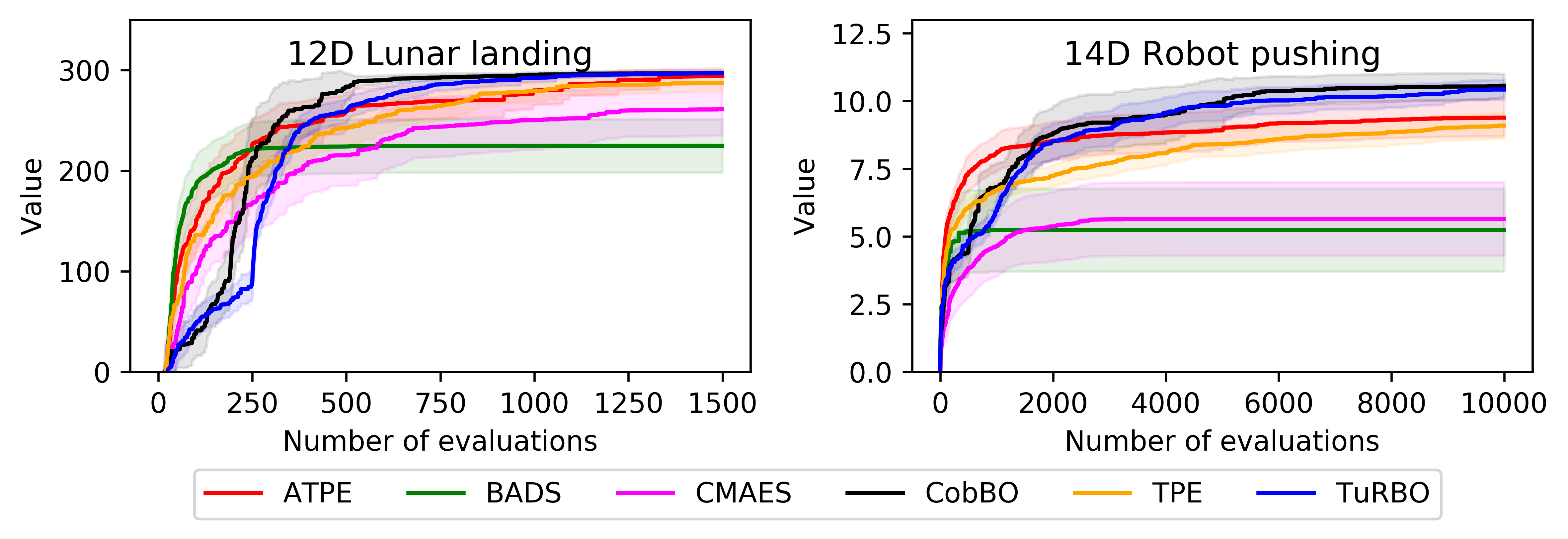}
  \vspace{-2mm}
\end{center}
\vspace{-1mm}
  \caption{Performance over the lunar landing (left) and robot pushing (right) problems}
  \label{fig:lunar-robot}
\end{figure*}

 \vspace{0.25cm}
\noindent \emph{Lunar landing (maximization):}
This controller learning problem ($12$ dimensions) is provided by the OpenAI gym and evaluated in~\cite{turbo2019}.
%The controller of a lunar lander decides whether or not to fire the booster engine and the firing direction during landing,   
%based on the current status of the lander in each frame. 
%The average performance of the controller is evaluated by simulations over %a fixed constant set of 
%50 randomly generated terrains and initial states. 
Each algorithm has 50 initial points and a budget of $1,500$ trials. 
TuRBO is configured with 5 trust regions and a batch size of 50 as in~\cite{turbo2019}.   
Fig.~\ref{fig:lunar-robot} shows that, among the $30$ independent tests, CobBO quickly exceeds $300$ along some good sample paths, outperforming other algorithms. 

 %\vspace{0.25cm}
\noindent \emph{Robot pushing (maximization):}
This control problem (14 dimensions) is introduced in~\cite{wang18aistats} and extensively tested in~\cite{turbo2019}.  We follow the setting in~\cite{turbo2019}, where TuRBO is configured with a batch size of 50 and 15 trust regions where each has 30 initial points.  
We exclude REMBO that takes too long per run (more than $24$ hours).  
Each experiment has a budget of $10,000$ evaluations.
On average CobBO exceeds 10.0 within 5,500 trials, while TuRBO requires about 7,000, 
as shown in Fig. ~\ref{fig:lunar-robot}.
TPE and ATPE converge to around 9.0, outperforming BADS and CEM-ES with large margins. 
The latter two exhibit large variations and get stuck at local optima.

% CobBO finds the best results for the robot pushing problem, 
% slightly outperforming TuRBO, as shown in Fig. ~\ref{fig:control-additive}. 
% Both TPE and ATPE  are less competitive but still outperform BADS and CMA-ES with large margins. 
% The latter two algorithms show large variations and get stuck in suboptima at very early stages.

% \vspace{0.25cm}
 \begin{figure*}[htb]\vspace{-0mm}
  \centering
  \includegraphics[width=1.0\linewidth,height=!]{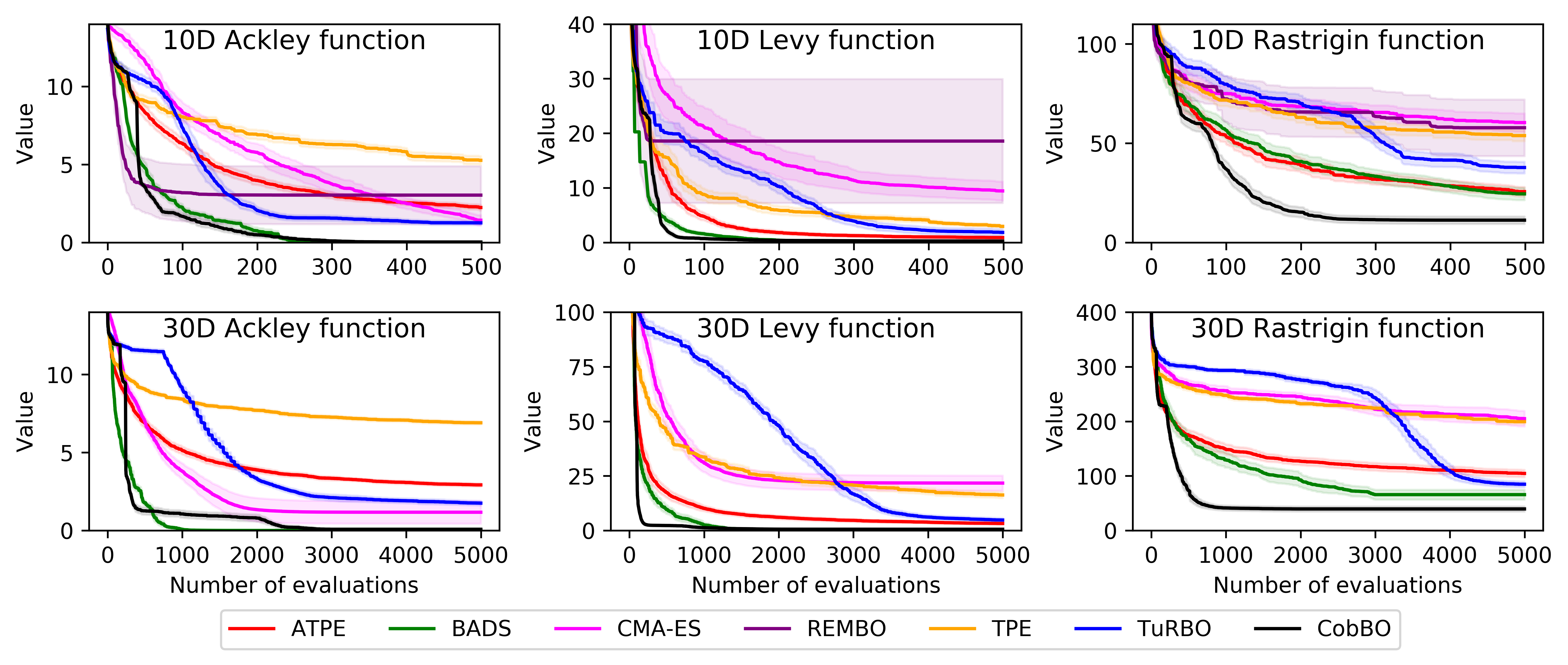}\vspace{-1mm}
  \caption{Performance on 10D (top) and 30D (bottom) synthetic black-box functions: Ackley (left), Levy (middle) and Rastrigin (right)}\vspace{-0.5mm}
  \label{fig:synthetic}
\end{figure*}
\noindent \emph{Classic synthetic black-box functions (minimization):}
Three popular synthetic functions ($10$ and $30$ dimensions) are chosen, including Ackley over $[-5, 10]^{10}$ and $[-5, 10]^{30}$, Levy over both $[-5, 10]^{10}$ and $[-5, 10]^{30}$, and Rastrigin over both $[-3, 4]^{10}$ and $[-3, 4]^{30}$.
%, and Hartmann(6D) with domain $[0, 1]^{6}$
%Each experiment has a budget of $500$ evaluations. 
TuRBO is configured identically the same as in~\cite{turbo2019}, with a batch size of $10$ and $5$ concurrent trust regions where each has $10$ initial points. 
%\niv{What does it mean "$5$ trust regions" ?}
The other algorithms use $20$ initial points. 
The results are shown in Fig.~\ref{fig:synthetic}. CobBO shows competitive or better performance for all of these problems.
It finds the global optima on Ackley and Levy, and clearly outperforms the other algorithms for the difficult Rastrigin function. 
Notably, BADS is more suitable for low dimensions, as commented in~\cite{luigi2017}, which performs close to CobBO except on Rastrigin. 
TuRBO performs better than TPE and worse than BADS. ATPE outperforms TPE. % and is close to CobBO on Levy.
CMA-ES eventually catches up with TPE, ATPE and REMBO on Ackley.
For $10$ dimensions, REMBO appears unstable with large variations and is trapped at local optima. 
For $30$ dimensions, REMBO is excluded as it takes too long to finish; see Section~\ref{ss:alebo}.
%(more than 24 hours per experiment in this case). 

\subsubsection{Comparison to REMBO and ALEBO}\label{ss:alebo}
REMBO~\cite{ziyuw2016} and ALEBO~\cite{letham2020} are designed for high-dimensional (large $D$) problems with low intrinsic dimensions (small $d$),  which essentially assumes that the function does not change along certain directions.
They do not necessarily perform well for problems without redundant dimensions, as shown by the following experiments with $D=d$. 
%To demonstrate this, we test using experiment with $D=d$.

First, we compare REMBO and CobBO using Ackley 200D with $4000$ iterations and $50$ initial points. Even though $D=d=200$ in this case, we treat REMBO as if the effective dimension were $d=20$, similar to CobBO's subspaces with an average size about $15$. 
REMBO and CobBO reach the mean best values of $15.1$ and $3.8$, respectively, running for $31.2$ and $3.4$ hours, respectively. 
  This shows that CobBO could outperform REMBO by a large margin for problems without redundant dimensions. In addition, CobBO requires about $10\%$ of the computation time of REMBO for this experiment, 
  which demonstrates the advantage of the two-stage kernels in reducing the computation time. 

Next, we compare with ALEBO, which has demonstrated great performance for problems with large $D$ but small $d$ in~\cite{letham2020}. Through extensive experiments we find that ALEBO works only when the underlying effective dimension satisfies $d\leq 20$. Otherwise, the algorithm suffers from the same curse of dimensionality as vanilla BO algorithms do, since the subproblem in the embedding space of $d$ dimensions is also challenging for large $d$. 

\begin{figure}[ht]
    \centering
    \vspace{-0.0cm}
    \includegraphics[width=1.0\textwidth, height=!]{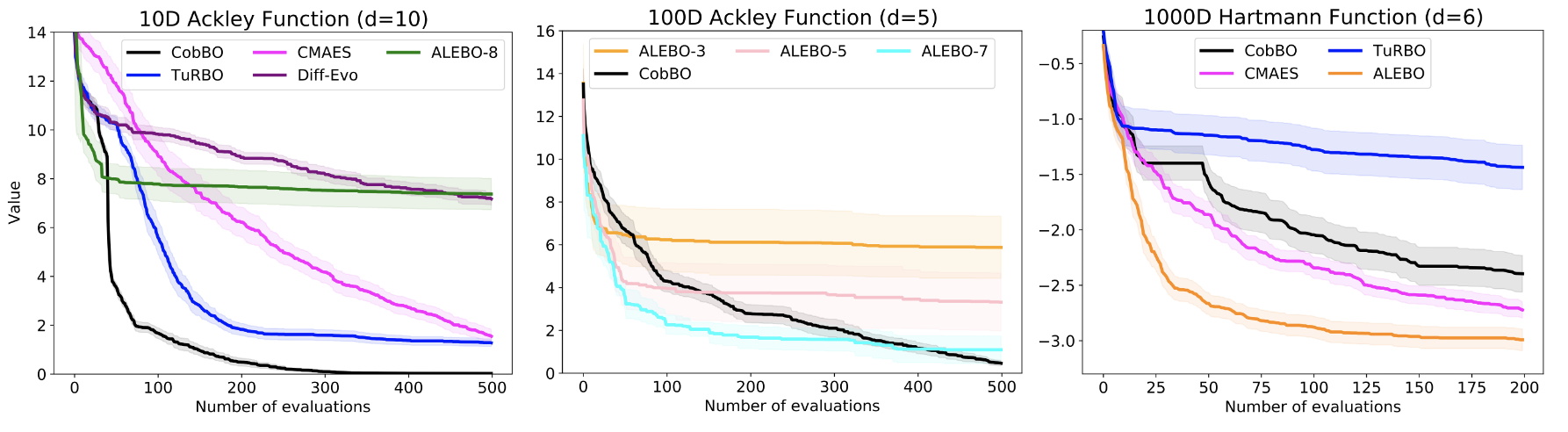} %ackley10-alebo.pdf}
  \caption{Experiments with $D=10, 100, 1000$ spaces of small effective dimensions $d=10, 5, 6$, respectively}
  \label{fig:alebo}
      \vspace{-0.0cm}
\end{figure}

To this end, we design three different experiments.  First, we study the general problems for $D=d$. Since ALEBO has performance issues for large $d$, we test Ackley ($D=d=10$). 
As ALEBO requires $d<D$, we treat it as if $d=8$ (ALEBO-8). In this case, ALEBO does not show good performance and is outperformed by CobBO, TurBO and CMAES, as shown in Fig.~\ref{fig:alebo} (left).  Second, we test Ackley ($D=100, d=5$). In reality, we do not know the effective dimension $d$. Therefore, we teat it as if $d=3, 5, 7$ to obtain ALEBO-3, ALEBO-5 and ALEBO-7, respectively.  Although this problem indeed has a very small $d=5$, CobBO can still perform well compared to ALEBO, as shown in Fig.~\ref{fig:alebo} (middle). The third experiment is using exactly the same setting as in~\cite{letham2020} for Hartmann6 with $D=1000$ and $d=6$.  
As shown in Fig.~\ref{fig:alebo} (right),  ALEBO outperforms CobBO, 
since CobBO is not designed for a function with a very high dimension $D=1000$ and a very low effective dimension $d=6$.  The reason is because CobBO relies on selection of subspaces of an average dimension $15$, which cannot easily cover the optima in a high dimensional space $D \geq 1000$.  In this case, after projecting the original function into a low $d$ dimensional embedding space, CobBO can be applied to solve the subproblem when $d$ is still considered to be too large, e.g., $d > 20$.

\subsubsection{Comparison to LineBO}\label{ss:linebo}
\begin{figure}[ht]
    \centering
    \vspace{-0.2cm}
        \includegraphics[width=0.75\textwidth, height=!]{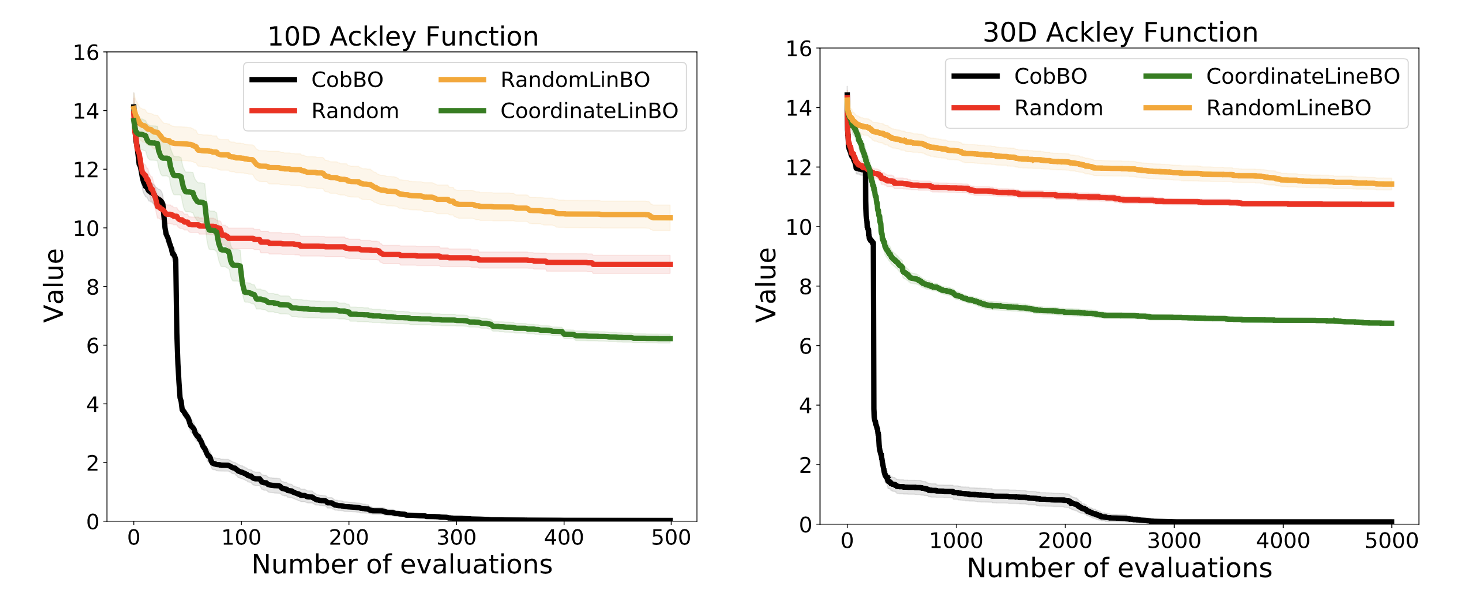} %linebo-10d}
        \vspace{-0.2cm}
   \caption{CobBO outperforming different variants of LineBO}
   \label{fig:linebo}    \vspace{-0.1cm}
\end{figure}

%  Although sharing some common basic ideas, LineBO~\cite{linebo} reduces the acquisition maximization cost by restricting on a line but does not reduce the expensive computational costs of the GP regression in the full space.
 Although sharing some common basic ideas, LineBO~\cite{linebo} reduces the acquisition maximization cost by restricting on a line, but it does not address the computational issue of the GP regression in the full space by using a single kernel, i.e., the first stage of CobBO.
 In addition, it is difficult to find a good direction to form the line space at each iteration, since searching for the optima in a high dimensional space on a random line is not computationally efficient.
 %which greatly impacts the efficiency and solution quality.  
% For the running time, we compare LineBO and CobBO using Ackley 100D 
% and 200D with $4,000$ trials and $50$ initial points, which take XX and YY hours to finish, respectively.
 Fig.~\ref{fig:linebo} shows that LineBO is significantly outperformed by CobBO using a typical example, e.g., Ackley, with $D=10, 30$. 
For $D=10$ with a query budget of $500$,  CobBO almost reaches the optimal solution $0.0$ while LineBO (CoordinateLineBO) only obtains 6.2.  
For $D=30$ with a query budget of $5000$, 
CobBO reaches $0.12$ and LineBO (CoordinateLineBO) only obtains $7.6$. In both cases, RandomLineBO performs even worse than random search.

% CobBO takes $10817$ seconds to reach 0.12 and LineBO takes $7500$ seconds to reach 7.6 (CoordinateLineBO). In both cases, RandomLineBO performs even worse than random search.  
%To further compare the running time in high dimensions, we test Ackley $D=100,200$, and the running times are XX and YY seconds.  

% \begin{figure}[ht]
%     \centering
%     \includegraphics[width=0.45\textwidth, height=!]{supplementary/hart1000.pdf}
%   \caption{Performance on Hartmann6 ($D=1000$, $d=6$)}
%   \label{fig:hart6}
% \end{figure}

% \begin{figure}[ht]
%     \centering
%     \includegraphics[width=0.32\textwidth, height=!]{supplementary/ackley10-alebo-2.pdf}
%     \includegraphics[width=0.32\textwidth, height=!]{supplementary/ackley10-alebo-4.pdf}
%     \includegraphics[width=0.32\textwidth, height=!]{supplementary/ackley10-alebo-8.pdf}
%   \caption{Compare ALEBO and CobBO on Ackley(10D)}
%   \label{fig:alebo}
% \end{figure}
%However, it takes ALEBO $12$ hours on average and only $10$ minutes for CobBO to finish the experiment.

% Regarding the computation times, it takes $6$ to $12$ hours for ALEBO and only $3$ minutes for CobBO to finish $500$ queries for each experiment on our testbed for the second case. 

% \vspace{-3mm}
\section{Conclusion}
% \vspace{-1.5mm}
CobBO is a variant of coordinate ascent tailored for Bayesian optimization with a stopping rule to switch coordinate subspaces. The sampling policy of subspaces is proven to have tight regret bounds with respect to the best subspace in hindsight. Combining the projection on random subspaces with a two-stage kernels for function value interpolation and GP regression, we provide a practical Bayesian optimization method of affordable computational costs in high dimensions.
%with the performance further improved by auxiliary features, %including trust regions alternation and data filtering.
Empirically, CobBO consistently finds comparable or better solutions with reduced trial complexity in comparison with the state-of-the-art methods across a variety of benchmarks.
% \section{Proofs}\label{sec:regret_analysis} 
% %This part contains the detailed proofs for the regret analysis. 
% \input{regret/regret_combinatorial}
% \input{regret/regret_analysis} 
% % \input{regret/regret_analysis_linear_T}
% \input{regret/regret_backoff}

\bibliography{references}
\bibliographystyle{icml2021}

\appendix

% \documentclass{article}

% \usepackage{algorithm2e}
% \usepackage{array, tabularx, caption, boldline}
% \usepackage{graphicx}
% \usepackage{makecell}
% \usepackage{url, amsmath}
% \usepackage{icml2021}

% \begin{document}
% \onecolumn
% \icmltitle{
% CobBO: Coordinate Backoff Bayesian Optimization \\
% Supplementary Materials}
\clearpage
\appendix
\begin{center}
    \huge
    \textbf{Supplementary Material}
\end{center}

\section{Proofs}\label{sec:regret_analysis} 
%This part contains the detailed proofs for the regret analysis. 
In this section we provide proofs for the theorems in Section~\ref{ss:analysis}.
To make non-negative temporal losses, we modify the losses in Eq.~(\ref{eq:alpha_beta_loss}) to be non-negative by adding the same constant $\log(\tilde{\alpha})$,
\begin{align} %\label{eq:alpha_beta_loss_shifted}
    \tilde{\ell}_{t,i} =
    \begin{cases}
		0 & \text{if } i \in C_t \text{ and } y_t > M_{t-1} \\
		\log(\tilde{\alpha}\tilde{\beta}) & \text{if } i \in C_t \text{ and } y_t \leq M_{t-1} \\
        \log(\tilde{\alpha}) & \text{if } i \notin C_t. \nonumber
    \end{cases}
\end{align}
This modification does not change the resulted distribution $\pi_t$ induced over the coordinates as it is invariant to shifts of the losses, $\pi_{t,i} = w_{t,i}/W_t$, 
\begin{align}
\notag
\pi_{t,i} = \frac{e^{-\eta\sum_{\tau=1}^t\tilde{\ell}_{\tau,i}}}{\sum_{j=1}^D e^{-\eta\sum_{\tau=1}^t\tilde{\ell}_{\tau,j}}} 
&=
\frac{e^{-\eta\sum_{\tau=1}^t(\ell_{\tau,i}+\log(\tilde{\alpha}))}}{\sum_{j=1}^D e^{-\eta\sum_{\tau=1}^t(\ell_{\tau,j}+\log(\tilde{\alpha}))}}
\\ &=
\frac{e^{-\eta t\log(\tilde{\alpha})}e^{-\eta\sum_{\tau=1}^t\ell_{\tau,i}}}{e^{-\eta t\log(\tilde{\alpha})}\sum_{j=1}^D e^{-\eta\sum_{\tau=1}^t\ell_{\tau,j}}}
=
\frac{e^{-\eta\sum_{\tau=1}^t\ell_{\tau,i}}}{\sum_{j=1}^D e^{-\eta\sum_{\tau=1}^t\ell_{\tau,j}}}. \nonumber
\end{align}
Thus, $\tilde{\pi}_{t,i}$ and $\hat{\pi}_{t,i}$ introduced in Sections~\ref{sec:comb_regret_analysis}~and~ \ref{sec:regret_analysis_without_replacement} remain unchanged as well. For simplicity we refer to $\tilde{\ell}$ as $\ell$ throughout this section.

\subsection{Regret analysis for sampling from the combinatorial space of coordinate blocks} \label{sec:comb_regret_analysis}

The probability $\tilde{\pi}_{t,\mathcal{I}_t}$ of selecting a certain coordinate block $\mathcal{I}_t\subset \mathcal{I}=\{1,\cdots,D\}$ of size $|\mathcal{I}_t|=c\in\mathcal{C}$ follows sampling according to $\pi_t$ such that
\begin{align}\label{eq:block_prob}
    \tilde{w}_{t,\mathcal{I}_t} = 
    \prod_{i\in \mathcal{I}_t}w_{t,i}^{\frac{1}{|\mathcal{I}_t|}}
    \; , \quad
    \tilde{W}_t = \sum_{c\in\mathcal{C}}\sum_{\mathcal{I}_t\in\mathcal{S}_c}\tilde{w}_{t,\mathcal{I}_t}
    \; , \quad
    \tilde{\pi}_{t,\mathcal{I}_t} = \frac{\tilde{w}_{t,\mathcal{I}_t}}{\tilde{W}_t}
    \quad \forall \mathcal{I}_t \in \bigcup_{c\in\mathcal{C}}\mathcal{S}_c
\end{align}
with
\begin{equation} \label{eq:prob_sum_to_1_2}
    \sum_{c\in\mathcal{C}}\sum_{\mathcal{I}_t\in\mathcal{S}_c}\tilde{\pi}_{t,\mathcal{I}_t} =1.
\end{equation}

% \textbf{Proof} 
%\subsubsection*{Proof of Lemma~\ref{lem:regret_comb}}
\begin{lemma} \label{lem:regret_comb}
For $\eta >0$ and non-negative losses $\ell_{t,i}\geq 0$ the update rule in (\ref{eq:multiplicative_weight_update}) satisfies for any block of coordinates $\mathcal{I}^*$:
\begin{align}\label{eq:lemma_1_2}
 \sum_{t=1}^T\sum_{c\in\mathcal{C}}\sum_{\mathcal{I}_t\in \mathcal{S}_c} & \tilde{\pi}_{t,\mathcal{I}_t}\cdot\frac{1}{|\mathcal{I}_t|} \sum_{i\in\mathcal{I}_t}\ell_{t,i} 
 -\sum_{t=1}^T\frac{1}{|\mathcal{I}^*|}\sum_{i\in \mathcal{I}^*}\ell_{t,i}
\leq   \nonumber\\
 & 
\eta\sum_{t=1}^T\sum_{c\in\mathcal{C}}\sum_{\mathcal{I}_t\in \mathcal{S}_c}\tilde{\pi}_{t,\mathcal{I}_t}\cdot \left(\frac{1}{|\mathcal{I}_t|}\sum_{i\in\mathcal{I}_t}\ell_{t,i}\right)^2 + \frac{D\log(D)}{\eta}.
\end{align}
\end{lemma}

%\begin{proof}
\noindent \textit{Proof}: 
Set
\begin{equation} \label{eq:init_2}
    \tilde{w}_{0,\mathcal{I}_t} = 1 \quad \forall \mathcal{I}_t \in \bigcup_{c\in\mathcal{C}}\mathcal{S}_c 
\end{equation}
Thus,
\begin{align}
    \tilde{W}_{t+1} &= 
    \sum_{c\in\mathcal{C}}\sum_{\mathcal{I}_t\in\mathcal{S}_c}\tilde{w}_{t+1,\mathcal{I}_t}
    =
    \sum_{c\in\mathcal{C}}\sum_{\mathcal{I}_t\in\mathcal{S}_c}\prod_{i\in \mathcal{I}_t}w_{t+1,i}^{\frac{1}{|\mathcal{I}_t|}} \nonumber
    \\ & =
    \sum_{c\in\mathcal{C}}\sum_{\mathcal{I}_t\in\mathcal{S}_c}\prod_{i\in \mathcal{I}_t}w_{t,i}^{\frac{1}{|\mathcal{I}_t|}}e^{-\frac{\eta}{|\mathcal{I}_t|}\ell_{t,i}}
    =
    \sum_{c\in\mathcal{C}}\sum_{\mathcal{I}_t\in\mathcal{S}_c}\prod_{i\in \mathcal{I}_t}w_{t,i}^{\frac{1}{|\mathcal{I}_t|}}\cdot e^{-\frac{\eta}{|\mathcal{I}_t|}\sum_{i\in\mathcal{I}_t}\ell_{t,i}} \nonumber\\
    & = 
    \sum_{c\in\mathcal{C}}\sum_{\mathcal{I}_t\in\mathcal{S}_c}\tilde{w}_{t,\mathcal{I}_t}\cdot e^{-\frac{\eta}{|\mathcal{I}_t|}\sum_{i\in\mathcal{I}_t}\ell_{t,i}} \nonumber
    \\ & = \label{eq:back_to_pi_hat_2}
    \tilde{W}_t\sum_{c\in\mathcal{C}}\sum_{\mathcal{I}_t\in\mathcal{S}_c}\tilde{\pi}_{t,\mathcal{I}_t}\cdot e^{-\frac{\eta}{|\mathcal{I}_t|}\sum_{i\in\mathcal{I}_t}\ell_{t,i}}\\
    &\leq \label{eq:exp_ineq_1_2}
    \tilde{W}_t\sum_{c\in\mathcal{C}}\sum_{\mathcal{I}_t\in \mathcal{S}_c}\tilde{\pi}_{t,\mathcal{I}_t}\left(1-\frac{\eta}{|\mathcal{I}_t|}\sum_{i\in\mathcal{I}_t}\ell_{t,i}+\eta^2\left(\frac{1}{|\mathcal{I}_t|}\sum_{i\in\mathcal{I}_t}\ell_{t,i}\right)^2\right)\\
    %         \end{align}
    % \begin{align}
    % % \\ 
    &\leq \label{eq:pi_hat_sums_to_1_2}
    \tilde{W}_t\left(1+
    \sum_{c\in\mathcal{C}}\left(\sum_{\mathcal{I}_t\in \mathcal{S}_c} \eta^2\tilde{\pi}_{t,\mathcal{I}_t}\left(\frac{1}{|\mathcal{I}_t|}\sum_{i\in\mathcal{I}_t}\ell_{t,i}\right)^2 -\frac{\eta}{|\mathcal{I}_t|}\tilde{\pi}_{t,\mathcal{I}_t}\sum_{i\in\mathcal{I}_t}\ell_{t,i}\right)
    \right)
    \\&\leq \label{eq:exp_ineq_2_2}
    \tilde{W}_t e^{
    \sum_{c\in\mathcal{C}}\cdot\left(\sum_{\mathcal{I}_t\in \mathcal{S}_c} \eta^2\tilde{\pi}_{t,\mathcal{I}_t}\left(\frac{1}{|\mathcal{I}_t|}\sum_{i\in\mathcal{I}_t}\ell_{t,i}\right)^2 -\frac{\eta}{|\mathcal{I}_t|}\tilde{\pi}_{t,\mathcal{I}_t}\sum_{i\in\mathcal{I}_t}\ell_{t,i}\right),
    } 
\end{align}
where (\ref{eq:back_to_pi_hat_2}) follows from (\ref{eq:block_prob}), (\ref{eq:exp_ineq_1_2}) holds since $e^{-x}\leq 1-x+x^2$ for $x\geq 0$, (\ref{eq:pi_hat_sums_to_1_2}) holds due to Eq.~(\ref{eq:prob_sum_to_1_2}) and (\ref{eq:exp_ineq_2_2}) holds since $1+x\leq e^x$.
% \begin{itemize}
%     \item (\ref{eq:back_to_pi_hat_2}) follows from (\ref{eq:block_prob}).
%     \item (\ref{eq:exp_ineq_1_2}) holds since $e^{-x}\leq 1-x+x^2$ for $x\geq 0$.
%     \item (\ref{eq:pi_hat_sums_to_1_2}) holds due to Eq.~\ref{eq:prob_sum_to_1_2}.
%     \item (\ref{eq:exp_ineq_2_2}) holds since $1+x\leq e^x$.
% \end{itemize}

Due to Eq.~(\ref{eq:init_2}),  we have,
\begin{align} \label{eq:aggergate_block_losses}
  \tilde{w}_{t,\mathcal{I}_t} 
  = 
  \prod_{i\in \mathcal{I}_t}w_{t,i}^{\frac{1}{|\mathcal{I}_t|}} 
  =
  \prod_{i\in \mathcal{I}_t}w_{0,i}^{\frac{1}{|\mathcal{I}_t|}} e^{-\frac{\eta}{|\mathcal{I}_t|}\sum_{t=1}^T\ell_{t,i}}
  =  e^{-\frac{\eta}{\mathcal{I}_t}\sum_{t=1}^T\sum_{i\in \mathcal{I}_t}\ell_{t,i}}.
\end{align}
And,
\begin{equation}\label{eq:big_init_2}
   W_0=\sum_{c\in\mathcal{C}}\sum_{\mathcal{I}_t\in\mathcal{S}_c}\tilde{w}_{0,\mathcal{I}_t}
   =\sum_{c\in\mathcal{C}}\sum_{\mathcal{I}_t\in\mathcal{S}_c}1
   =\sum_{c\in\mathcal{C}}|\mathcal{S}_c|=
   \sum_{c\in\mathcal{C}}{D \choose c} \leq (D!)^{|\mathcal{C}|}.
\end{equation}
Given that the weight of a certain coordinate block $\mathcal{I}^*$ is less than the total sum of all weights, together with Eq.~(\ref{eq:exp_ineq_2_2}),~(\ref{eq:init_2}) and~(\ref{eq:big_init_2}) we have
\begin{align}
   e^{-\frac{\eta}{|\mathcal{I}^*|}\sum_{t=1}^T\sum_{i\in \mathcal{I}^*}\ell_{t,i}}
   &=  
   \tilde{w}_{t,\mathcal{I}^*} \leq \tilde{W}_T 
   \notag \\ 
   &%\hspace{1cm}
   \leq 
   (D!)^{|\mathcal{C}|} e^{
   \sum_{t=1}^T\sum_{c\in\mathcal{C}}\cdot\left(\sum_{\mathcal{I}_t\in \mathcal{S}_c} \eta^2\tilde{\pi}_{t,\mathcal{I}_t}\left(\frac{1}{|\mathcal{I}_t|}\sum_{i\in\mathcal{I}_t}\ell_{t,i}\right)^2 -\frac{\eta}{|\mathcal{I}_t|}\tilde{\pi}_{t,\mathcal{I}_t}\sum_{i\in\mathcal{I}_t}\ell_{t,i}\right)
  }. \nonumber
\end{align}
Taking the $\log$ of both sides, we have
\begin{align} %\label{eq:recursive_ineq_2}
-\eta\sum_{t=1}^T\frac{1}{|\mathcal{I}^*|}\sum_{i\in \mathcal{I}^*}\ell_{t,i}
  \leq &
  \sum_{t=1}^T\sum_{c\in\mathcal{C}}\cdot\left(\sum_{\mathcal{I}_t\in \mathcal{S}_c} \eta^2\tilde{\pi}_{t,\mathcal{I}_t}\left(\frac{1}{|\mathcal{I}_t|}\sum_{i\in\mathcal{I}_t}\ell_{t,i}\right)^2 -\frac{\eta}{|\mathcal{I}_t|}\tilde{\pi}_{t,\mathcal{I}_t}\sum_{i\in\mathcal{I}_t}\ell_{t,i}\right) 
% \notag \\& 
  + |\mathcal{C}|\log(D!), \nonumber
\end{align}
which, using $D!\leq D^D$, finishes the proof.
%\end{proof}

%  \textbf{Proof} 
\noindent \textbf{Proof of Theorem~\ref{theo:regret_comb}}:
%\begin{proof}
%\noindent \textbf{Proof}: 
Since $\ell_{t,i} \leq \log(\tilde{\alpha}\tilde{\beta})$, then 
 \begin{align}
     \left(\frac{1}{|\mathcal{I}_t|}\sum_{i\in\mathcal{I}_t}\ell_{t,i}\right)^2
     \leq
     \left(\frac{1}{|\mathcal{I}_t|}\sum_{i\in\mathcal{I}_t}\log(\tilde{\alpha}\tilde{\beta})\right)^2
     \leq
     \log(\tilde{\alpha}\tilde{\beta})^2.  \nonumber
 \end{align}
 
 Thus, due to Eq.~(\ref{eq:prob_sum_to_1_2}), one has
 \begin{align}
    \sum_{c\in\mathcal{C}}\sum_{\mathcal{I}_t\in \mathcal{S}_c}\tilde{\pi}_{t,\mathcal{I}_t}\cdot \left(\frac{1}{|\mathcal{I}_t|}\sum_{i\in\mathcal{I}_t}\ell_{t,i}\right)^2
     \leq
     \sum_{c\in\mathcal{C}}\sum_{\mathcal{I}_t\in \mathcal{S}_c}\tilde{\pi}_{t,\mathcal{I}_t}\log(\tilde{\alpha}\tilde{\beta})^2
     =
     \log(\tilde{\alpha}\tilde{\beta})^2. \nonumber
 \end{align}
 Setting $\eta=\frac{1}{\log(\tilde{\alpha}\tilde{\beta})}\sqrt{\frac{|\mathcal{C}|D\log(D)}{T}}$ in Eq.~(\ref{eq:lemma_1_2}) yields
 \begin{equation}
     Regret_t \leq \eta T \log(\tilde{\alpha}\tilde{\beta})^2 + \frac{|\mathcal{C}|D\log(D)}{\eta}
    = 
     2 \log(\tilde{\alpha}\tilde{\beta})\sqrt{T|\mathcal{C}|D\log(D)}.
 \end{equation}
%\end{proof}

\subsection{Regret analysis for sampling coordinates without replacement} \label{sec:regret_analysis_without_replacement}
Denote by $p_c$ the probability of choosing a certain block size $c\in\mathcal{C}$, such that $p_c> 0$ and $\sum_{c\in\mathcal{C}}p_c=1$, e.g., for a uniform sampling of the block size $p_c=1/|\mathcal{C}|$ for all $c\in\mathcal{C}$.

The probability $\hat{\pi}_{t,\mathcal{I}_t}$ of selecting a certain coordinate block $\mathcal{I}_t\subset \mathcal{I}=\{1,\cdots,D\}$ of size $|\mathcal{I}_t|=c\in\mathcal{C}$ follows sampling according to $\pi_t$ (Eq.~(\ref{eq:multiplicative_update})) without replacement, such that,
\begin{align}
    \hat{\pi}_{t,\mathcal{I}_t} 
    &= \sum_{p\in perm(\mathcal{I}_t)} \prod_{k\in p}{\frac{\pi_{t,k}}{1-\sum_{j\in p_{1:k}}\pi_{t,j}}} \nonumber
    \\&= \label{eq:common_numerator}
    \left(\prod_{i\in \mathcal{I}_t}\pi_{t,i}\right) \cdot \left(\sum_{p\in perm(\mathcal{I}_t)} \prod_{k\in p}\left(1-\sum_{j\in p_{1:k}}\pi_{t,j}\right)^{-1}\right)
    =
    \mathcal{P}(\mathcal{I}_t) \cdot \mathcal{R}(\mathcal{I}_t)
\end{align}
where $perm(\mathcal{I}_t)$ are all the permutations of the set $\mathcal{I}_t$ and $p_{1:k}$ are the first $k$ coordinates in the permutation $p$. Eq.~(\ref{eq:common_numerator}) holds due to the common numerator of all permutations where the left term $\mathcal{P}(\mathcal{I}_t)$ corresponds to the probability of sampling a subset of coordinates with replacement, and the right term $\mathcal{R}(\mathcal{I}_t)$ is associated with sampling without replacement. Of course, summing over all the possible blocks of size $c$ results $\sum_{\mathcal{I}_t\in\mathcal{S}_c}\hat{\pi}_{t,\mathcal{I}_t} = 1$ for all $c\in\mathcal{C}$.

Thus $\tilde{\pi}_{t,\mathcal{I}_t}=p_c\cdot\hat{\pi}_{t,\mathcal{I}_t}$ and the probability of sampling every block of coordinates of any size sum up to $1$ as well:
\begin{equation}\label{eq:prob_sum_to_1}
    \sum_{c\in\mathcal{C}}\sum_{\mathcal{I}_t\in\mathcal{S}_c}\tilde{\pi}_{t,\mathcal{I}_t} 
    =
    \sum_{c\in\mathcal{C}}p_c\sum_{\mathcal{I}_t\in\mathcal{S}_c}\hat{\pi}_{t,\mathcal{I}_t} 
    = 
    \sum_{c\in\mathcal{C}}p_c
    = 
    1
\end{equation}

% Denote the set of all possible coordinate blocks of all sizes $c\in\mathcal{C}$ by $\mathcal{S}$ and its size $|\mathcal{S}|=\prod_{c\in\mathcal C} {D \choose c}$.

% The expected cumulative loss following our update policy is
% $$L_T= \sum_{t=1}^T\sum_{\mathcal{I}_t\in \mathcal{S}}\hat{\pi}_{t,\mathcal{I}_t}\cdot\frac{1}{|\mathcal{I}_t|} \sum_{i\in\mathcal{I}_t}\ell_{t,i}$$

% Assume the best coordinate block is $\mathcal{I}^*$ and the corresponding cumulative loss:
% $$L_T^*= \sum_{t=1}^T L_{t,\mathcal{I}_t}=\sum_{t=1}^T\frac{1}{|\mathcal{I}_t|} \sum_{i\in\mathcal{I}_t}\ell_{t,i} = \sum_{t=1}^T\bar{\ell}_{t,\mathcal{I}_t}$$

% We hence aim at bounding the regret $Regret_T = L_T-L_T^*$ 
% For this purpose we bound the regret with respect to any arbitrary sequence of selected coordinate blocks.

% \textbf{Lemma 1} For $\eta >0$ and assume all losses $\ell_t,i \geq 0$ for all $t\in\{1,\cdots,T\}$ and $i\in \mathcal{I}$ the update rule in \ref{eq:multiplicative_weight_update} satisfies for any block of coordinates $\mathcal{I}^*$:
% \begin{equation}\label{eq:lemma_1}
% % Regret_T=L_T-L^*_T =
% \sum_{t=1}^T\sum_{\mathcal{I}_t\in \mathcal{S}}\hat{\pi}_{t,\mathcal{I}_t}\cdot\frac{1}{|\mathcal{I}_t|} \sum_{i\in\mathcal{I}_t}\ell_{t,i} - \sum_{t=1}^T \ell^*_t
% \leq \eta \sum_{t=1}^T\sum_{\mathcal{I}_t\in \mathcal{S}}\hat{\pi}_{t,\mathcal{I}_t}\cdot \left(\frac{1}{|\mathcal{I}_t|}\sum_{i\in\mathcal{I}_t}\ell_{t,i}\right)^2 + \frac{\log(D)}{\eta}
% \end{equation}

% \textbf{Proof} 
%\subsubsection{Proof of Lemma~\ref{lem:regret_without_replacement}}
\begin{lemma} \label{lem:regret_without_replacement}
 Sample a block size $c\in\mathcal{C}$ with probability $p_c>0$ and $c$ coordinates without replacement according to $\pi_t$. Assume $\mathcal{C}\supset\{1\}$, $\eta >0$ and non-negative losses $\ell_{t,i}\geq 0$. Then the update rule in (\ref{eq:multiplicative_weight_update}) satisfies for any block of coordinates $\mathcal{I}^*$:
\begin{align}\label{eq:lemma_1}
% Regret_T
% =L_T-L^*_T 
% =
\sum_{t=1}^T\sum_{c\in\mathcal{C}}p_c\sum_{\mathcal{I}_t\in \mathcal{S}_c}\hat{\pi}_{t,\mathcal{I}_t}\cdot&\frac{1}{|\mathcal{I}_t|} \sum_{i\in\mathcal{I}_t}\ell_{t,i} 
 -\sum_{t=1}^T\frac{1}{|\mathcal{I}^*|}\sum_{i\in \mathcal{I}^*}\ell_{t,i}
% \sum_{t=1}^T\sum_{\mathcal{I}_t\in \mathcal{S}}\hat{\pi}_{t,\mathcal{I}_t}\cdot\frac{1}{|\mathcal{I}_t|} \sum_{i\in\mathcal{I}_t}\ell_{t,i} - \sum_{t=1}^T \ell^*_t
%  \sum_{t=1}^T\sum_{c\in\mathcal{C}}p_c\sum_{\mathcal{I}_t\in \mathcal{S}_c}\hat{\pi}_{t,\mathcal{I}_t}\cdot\frac{1}{|\mathcal{I}_t|} \sum_{i\in\mathcal{I}_t}\ell_{t,i} - \sum_{t=1}^T \ell^*_t
\notag \\ & \leq 
\eta\sum_{t=1}^T\sum_{c\in\mathcal{C}}p_c\sum_{\mathcal{I}_t\in \mathcal{S}_c}\hat{\pi}_{t,\mathcal{I}_t}\cdot \left(\frac{1}{|\mathcal{I}_t|}\sum_{i\in\mathcal{I}_t}\ell_{t,i}\right)^2 + \frac{\log(D)}{\eta} -\frac{T\log(p_1)}{\eta}
\end{align}
\end{lemma}

%\begin{proof}
\noindent \textit{Proof}: 
Starting with a uniform distribution over the coordinates $w_{0,i}\equiv \frac{1}{D}$ such that $W_0 = 1$ and we have:
\begin{align}
    p_1\cdot W_{t+1} &= p_1\cdot\sum_{i\in \mathcal{I}} w_{t+1,i} 
    \nonumber\\ & \leq \label{eq:sum_of_products}
     \sum_{c\in\mathcal{C}}p_c\sum_{\mathcal{I}_t\in \mathcal{S}_c}\prod_{i\in\mathcal{I}_t}w_{t+1,i}
    \\&=
    W_t\sum_{c\in\mathcal{C}}p_c\sum_{\mathcal{I}_t\in \mathcal{S}_c}W_t^{-1}\prod_{i\in\mathcal{I}_t}w_{t,i}e^{-\eta\ell_{t,i}} 
    \nonumber \\&\leq \label{eq:sum_leq_1}
    W_t\sum_{c\in\mathcal{C}}p_c\sum_{\mathcal{I}_t\in \mathcal{S}_c}W_t^{-|\mathcal{I}_t|}\prod_{i\in\mathcal{I}_t}w_{t,i}e^{-\eta\ell_{t,i}} \cdot|perm(\mathcal{I}_t)|
    \\&=
    W_t \sum_{c\in\mathcal{C}}p_c\sum_{\mathcal{I}_t\in \mathcal{S}_c}\prod_{i\in\mathcal{I}_t}\frac{w_{t,i}}{W_t}e^{-\eta\ell_{t,i}}\cdot \sum_{p\in perm(\mathcal{I}_t)}1 
    \nonumber\\
    &=
    W_t\sum_{c\in\mathcal{C}}p_c\sum_{\mathcal{I}_t\in \mathcal{S}_c}\prod_{i\in\mathcal{I}_t}\pi_{t,i}e^{-\eta\ell_{t,i}}\cdot \sum_{p\in perm(\mathcal{I}_t)}\prod_{k\in p}1 
    \nonumber\\
    %      \end{align}
    %  \begin{align}
    &\leq 
    W_t\sum_{c\in\mathcal{C}}p_c\sum_{\mathcal{I}_t\in \mathcal{S}_c}e^{-\eta\sum_{i\in\mathcal{I}_t}\ell_{t,i}}\prod_{i\in\mathcal{I}_t}\pi_{t,i}\cdot\sum_{p\in perm(\mathcal{I}_t)}\prod_{k\in p}\left(1-\sum_{j\in p_{1:k}}\pi_{t,j}\right)^{-1}
    \nonumber
    \\&= \label{eq:back_to_pi_hat}
    W_t \sum_{c\in\mathcal{C}}p_c\sum_{\mathcal{I}_t\in \mathcal{S}_c}\hat{\pi}_{t,\mathcal{I}_t}e^{-\eta\sum_{i\in\mathcal{I}_t}\ell_{t,i}}
    \\&\leq 
    W_t \sum_{c\in\mathcal{C}}p_c\sum_{\mathcal{I}_t\in \mathcal{S}_c}\hat{\pi}_{t,\mathcal{I}_t}e^{-\frac{\eta}{|\mathcal{I}_t|}\sum_{i\in\mathcal{I}_t}\ell_{t,i}}
    \nonumber 
    \\&\leq \label{eq:exp_ineq}
    W_t\sum_{c\in\mathcal{C}}p_c\sum_{\mathcal{I}_t\in \mathcal{S}_c}\hat{\pi}_{t,\mathcal{I}_t}\left(1-\frac{\eta}{|\mathcal{I}_t|}\sum_{i\in\mathcal{I}_t}\ell_{t,i}+\eta^2\left(\frac{1}{|\mathcal{I}_t|}\sum_{i\in\mathcal{I}_t}\ell_{t,i}\right)^2\right)
    \\&\leq \label{eq:pi_hat_sums_to_1}
    W_t\left(1+
    \sum_{c\in\mathcal{C}}p_c\cdot\left(\sum_{\mathcal{I}_t\in \mathcal{S}_c} \eta^2\hat{\pi}_{t,\mathcal{I}_t}\left(\frac{1}{|\mathcal{I}_t|}\sum_{i\in\mathcal{I}_t}\ell_{t,i}\right)^2 -\frac{\eta}{|\mathcal{I}_t|}\hat{\pi}_{t,\mathcal{I}_t}\sum_{i\in\mathcal{I}_t}\ell_{t,i}\right)
    \right)
    \\&\leq \label{eq:exp_ineq_2}
    W_t e^{
    \sum_{c\in\mathcal{C}}p_c\cdot\left(\sum_{\mathcal{I}_t\in \mathcal{S}_c} \eta^2\hat{\pi}_{t,\mathcal{I}_t}\left(\frac{1}{|\mathcal{I}_t|}\sum_{i\in\mathcal{I}_t}\ell_{t,i}\right)^2 -\frac{\eta}{|\mathcal{I}_t|}\hat{\pi}_{t,\mathcal{I}_t}\sum_{i\in\mathcal{I}_t}\ell_{t,i}\right)
    } 
\end{align}
where
\begin{itemize}
     \item (\ref{eq:sum_of_products}) holds since $\mathcal{C}\supset\{1\}$ always contains a block size of $1$ and thus
    \begin{align*}
        \sum_{c\in\mathcal{C}}p_c\sum_{\mathcal{I}_t\in\mathcal{S}_c}\prod_{i\in\mathcal{I}_t}w_{t+1,i}
        &=
        p_1\sum_{\mathcal{I}_t\in\mathcal{S}_1}\prod_{i\in\mathcal{I}_t}w_{t+1,i}+\sum_{c\in\mathcal{C}\setminus \{1\}}p_c\sum_{\mathcal{I}_t\in\mathcal{S}_c}\prod_{i\in\mathcal{I}_t}w_{t+1,i}
        \\&=
        p_1\sum_{i\in \mathcal{I}} w_{t+1,i} + \sum_{c\in\mathcal{C}\setminus \{1\}}p_c\sum_{\mathcal{I}_t\in\mathcal{S}_c}\prod_{i\in\mathcal{I}_t}w_{t+1,i} 
       % \\ & 
       \geq 
        p_1\sum_{i\in \mathcal{I}} w_{t+1,i}
    \end{align*}
    \item (\ref{eq:sum_leq_1}) holds since $W_0=1$ and $W_t$ is monotonically non-increasing following the update rule (\ref{eq:multiplicative_weight_update}) with non-negative losses, thus $w_t\leq 1$ for all $t$; (\ref{eq:back_to_pi_hat}) follows from (\ref{eq:common_numerator}); (\ref{eq:exp_ineq}) holds since $e^{-x}\leq 1-x+x^2$ for $x\geq 0$; (\ref{eq:pi_hat_sums_to_1}) holds due to Eq.~\ref{eq:prob_sum_to_1}; (\ref{eq:exp_ineq_2}) holds since $1+x\leq e^x$.
\end{itemize}

Given that the sum of weights of a certain coordinate block $\mathcal{I}^*$ is less than the total sum of weights, together with Eq.~\ref{eq:exp_ineq_2}, $w_{0,i}\equiv\frac{1}{D}$ and $W_0=1$ we have
\begin{align}
   \frac{1}{D}\sum_{i\in \mathcal{I}^*} & e^{-\eta\sum_{t=1}^T\ell_{t,i}}
   = \sum_{i\in \mathcal{I}^*} w_{t,i} \leq W_T 
   \notag \\ &
   \leq 
    p_1^{-T} e^{
  \sum_{t=1}^T\sum_{c\in\mathcal{C}}p_c\cdot\left(\sum_{\mathcal{I}_t\in \mathcal{S}_c} \eta^2\hat{\pi}_{t,\mathcal{I}_t}\left(\frac{1}{|\mathcal{I}_t|}\sum_{i\in\mathcal{I}_t}\ell_{t,i}\right)^2 -\frac{\eta}{|\mathcal{I}_t|}\hat{\pi}_{t,\mathcal{I}_t}\sum_{i\in\mathcal{I}_t}\ell_{t,i}\right) 
  },\nonumber
  \end{align}
Taking the $\log$ of both sides, we have
\begin{align} \label{eq:recursive_ineq}
  \log&\left(\sum_{i\in \mathcal{I}^*}e^{-\eta\sum_{t=1}^T\ell_{t,i}}\right) -\log(D) 
  \notag \\ 
  & \leq
  \sum_{t=1}^T\sum_{c\in\mathcal{C}}p_c\cdot\left(\sum_{\mathcal{I}_t\in \mathcal{S}_c} \eta^2\hat{\pi}_{t,\mathcal{I}_t}\left(\frac{1}{|\mathcal{I}_t|}\sum_{i\in\mathcal{I}_t}\ell_{t,i}\right)^2 -\frac{\eta}{|\mathcal{I}_t|}\hat{\pi}_{t,\mathcal{I}_t}\sum_{i\in\mathcal{I}_t}\ell_{t,i}\right) - T\log(p_1)
  \end{align}
Following the same certain block, all the participating coordinates suffer the same loss 
 $\ell^*_t$ at every time step as follows from Eq.~\ref{eq:alpha_beta_loss}, hence
 \begin{align} 
%  \notag
     \log\left(\sum_{i\in \mathcal{I}^*}e^{-\eta\sum_{t=1}^T\ell_{t,i}}\right)
     =
     \log\left(\sum_{i\in \mathcal{I}^*}e^{-\eta\sum_{t=1}^T\ell^*_t}\right)
     &= 
     \notag
     \log\left(|\mathcal{I}^*|e^{-\eta\sum_{t=1}^T\ell^*_t}\right)
     \\&
     =  %\notag
     \log(|\mathcal{I}^*|) -\eta\sum_{t=1}^T\ell^*_t
     \geq %\label{eq:common_loss_ineq}
     -\eta\sum_{t=1}^T\ell^*_t, \nonumber
 \end{align}
 which, together with Eq.~(\ref{eq:recursive_ineq}),  yields
 %and (\ref{eq:common_loss_ineq}) yield
 \begin{align}
     -\eta\sum_{t=1}^T\ell^*_t &-\log(D) 
     \notag \\ & 
     \leq 
     \sum_{t=1}^T\sum_{c\in\mathcal{C}}p_c\cdot\left(\sum_{\mathcal{I}_t\in \mathcal{S}_c} \eta^2\hat{\pi}_{t,\mathcal{I}_t}\left(\frac{1}{|\mathcal{I}_t|}\sum_{i\in\mathcal{I}_t}\ell_{t,i}\right)^2 -\frac{\eta}{|\mathcal{I}_t|}\hat{\pi}_{t,\mathcal{I}_t}\sum_{i\in\mathcal{I}_t}\ell_{t,i}\right) -T\log(p_1), \nonumber
 \end{align}
which finishes the proof.
 %\end{proof}
 
%  \textbf{Theorem 1}  Apply the update rule in \ref{eq:multiplicative_weight_update}, with a modified $\eta=\log(\alpha\beta)^{-1}\sqrt{\frac{\log(D)}{T}}$, then:
%  \begin{equation*}
%      Regret_t = \mathcal{O}(\log(\alpha\beta)\sqrt{T\log(D)})
%  \end{equation*}
 
%  \textbf{Proof} 
\noindent \textbf{Proof of Theorem~\ref{theo:regret_without_replacement}}:
%\begin{proof}
Since $\ell_{t,i} \leq \log(\tilde{\alpha}\tilde{\beta})$ then
 \begin{align}
     \left(\frac{1}{|\mathcal{I}_t|}\sum_{i\in\mathcal{I}_t}\ell_{t,i}\right)^2
     \leq
     \left(\frac{1}{|\mathcal{I}_t|}\sum_{i\in\mathcal{I}_t}\log(\tilde{\alpha}\tilde{\beta})\right)^2
     \leq
     \log(\tilde{\alpha}\tilde{\beta})^2. \nonumber
 \end{align}
 Thus, due to Eq.~(\ref{eq:prob_sum_to_1}), we have
 \begin{align}
    \sum_{c\in\mathcal{C}}p_c\sum_{\mathcal{I}_t\in \mathcal{S}_c}\hat{\pi}_{t,\mathcal{I}_t}\cdot \left(\frac{1}{|\mathcal{I}_t|}\sum_{i\in\mathcal{I}_t}\ell_{t,i}\right)^2
     \leq
     \sum_{c\in\mathcal{C}}p_c\sum_{\mathcal{I}_t\in \mathcal{S}_c}\hat{\pi}_{t,\mathcal{I}_t}\log(\tilde{\alpha}\tilde{\beta})^2
     =
     \log(\tilde{\alpha}\tilde{\beta})^2. \nonumber
 \end{align}
Eq.~(\ref{eq:lemma_1}) reads
 \begin{equation} \label{eq:final_lemma_1}
     Regret_t \leq \eta T \log(\tilde{\alpha}\tilde{\beta})^2 + \frac{\log(D)}{\eta} - \frac{T\log(p_1)}{\eta}.
    %  = 
    %  2 \log(\tilde{\alpha}\tilde{\beta})\sqrt{T\log(D)}
 \end{equation}
Choosing $\eta\geq 1$, we have
 \begin{equation}
     Regret_t \leq \eta T \log(\tilde{\alpha}\tilde{\beta})^2 + \frac{\log(D)}{\eta} - \eta T\log(p_1) = 
     \eta T (\log(\tilde{\alpha}\tilde{\beta})^2 -\log(p_1)) + \frac{\log(D)}{\eta}. \nonumber
    %  = 
    %  2 \log(\tilde{\alpha}\tilde{\beta})\sqrt{T\log(D)}
 \end{equation}
 Thus setting $\eta=\sqrt{\frac{\log(D)}{T(\log(\tilde{\alpha}\tilde{\beta})^2 -\log(p_1))}}\geq 1$ finally we have
  \begin{equation}
     Regret_t 
     \leq 
     \mathcal{O}\left(\sqrt{
     (\log(\tilde{\alpha}\tilde{\beta})^2 -\log(p_1)) \cdot T\log(D)}
     \right). \nonumber
    %  2 \log(\tilde{\alpha}\tilde{\beta})\sqrt{T\log(D)}
 \end{equation}
 %\end{proof}
 
 \emph{Remark:} Note that the condition $\eta\geq 1$ can be replaced by setting an appropriate $p_1=\sqrt[T]{\epsilon}$ for $0<\epsilon\leq 1$. Thus Eq.~(\ref{eq:final_lemma_1}) reads
 \begin{equation}
     Regret_t 
     \leq 
     \eta T \log(\tilde{\alpha}\tilde{\beta})^2 + \frac{\log(D)-\log(\epsilon)}{\eta}.\nonumber
 \end{equation}
 Thus, setting $\eta=\frac{1}{\log(\tilde{\alpha}\tilde{\beta})}\sqrt{\frac{\log(D)-\log(\epsilon)}{T}}$ yields
     $Regret_t 
     \leq 
     \mathcal{O}\left(\log(\tilde{\alpha}\tilde{\beta})^{-1}\sqrt{
      T(\log(D)-\log(\epsilon))} \right)$.

\subsection{Regret analysis for consistent queries}\label{sec:regret_analysis_consistent_queries}
The regret analyses presented in Sections~\ref{sec:comb_regret_analysis}~and~\ref{sec:regret_analysis_without_replacement} hold when incorporating the consistent queries mentioned in section~\ref{ss:backoff} for an adapted settings.

Consider the update rule of Eq.~\ref{eq:multiplicative_weight_update} at each time step $t=1,\dots,T$ where the sampling of next coordinate blocks happens for $K\leq T$ time steps at $0=t_0<t_1<\dots<t_{k}<\dots<t_{K-1}<t_k=T$. Both $K$ and $\{t_k\}_{k=0}^{K-1}$ are unknown in advance and are revealed to the decision maker along the process. At each time $t_k$ a coordinate block is selected and fixed for the next $t_{k+1}-t_k$ steps. The effective losses incurred to the coordinates are the aggregation of all the temporal losses in this time interval $t\in[t_k, t_{k+1}-1]$, and thus $\bar{\ell}_{k,i} = \sum_{t=t_k}^{t_{k+1}-1} \ell_{t,i}$ where $\bar{\ell}_{k,i}\geq 0$ due to $\ell_{t,i}\geq 0$.

% This means that when the decision maker selects a coordinate block, it is consistently queried for some unknown time during which all the losses are aggregated to 

Since the update rule in Eq.~(\ref{eq:multiplicative_update}) is applied in every time step $t=1,\dots,T$, we effectively have
\begin{align}
    w_{k+1,i} = 
    w_{k,i}\prod_{t=t_k}^{t_{k+1}-1} e^{-\eta \ell_{t,i}} 
    = 
    w_{k,i} e^{-\eta \sum_{t=t_k}^{t_{k+1}-1}\ell_{t,i}} 
    =
    w_{k,i} e^{-\eta \bar{\ell}_{k,i}}. \nonumber
\end{align}

Define the stopping rule mentioned in section~\ref{ss:backoff} such that the number of consistent queries in a subspace does not cross $\tau\in[1,2,\dots,T]$, such that $
   t_{k+1}  - t_{k} \leq \tau$ for all $k=0,\dots,K-1$ 
% \begin{equation}
%   t_{k+1}  - t_{k} \leq \tau \quad \forall k=0,\dots,K-1 \nonumber
% \end{equation}
and thus $\bar{\ell}_{k,i}\leq \tau \log(\tilde{\alpha}\tilde{\beta})$ since $\ell_{t,i}\leq \log(\tilde{\alpha}\tilde{\beta})$.

Hence, all the results hold by replacing $T$ with $K$ and $\log(\tilde{\alpha}\tilde{\beta})$ with $\tau \log(\tilde{\alpha}\tilde{\beta})$.

\section{More on implementation and ablation studies} %IMPLEMENTATION}
The proposed CobBO algorithm is implemented in Python~3.  The source code and the original log files of all the experiments are attached for review. 
%and is publicly released online. 
%\subsection{Logs of experiments}
%The original log files of all the experiments are attached for the review. 
% The specifications of the testbed are as follows: CPU: Intel(R) Xeon(R) CPU E5-2682 v4 2.50GHz, Memory: 32GB, GPU: NVIDIA Tesla P100 PCIe 16GB.
The code has been utilized for various complex real-world applications and handles many corner cases (hence the error fallbacks). For example, a parameter ``smooth'' of Scipy RBF (kernel=multiquadric, default=0.0) is increased by 0.02 upon
``try catch''  numerical issues  of ill conditioning.

\subsection{Escaping trapped local optima}\label{ss:escaping}
\begin{figure}[!htb]%
  \centering
  \vspace{-0.3cm}
  \includegraphics[width=0.5
  \linewidth,height=!]{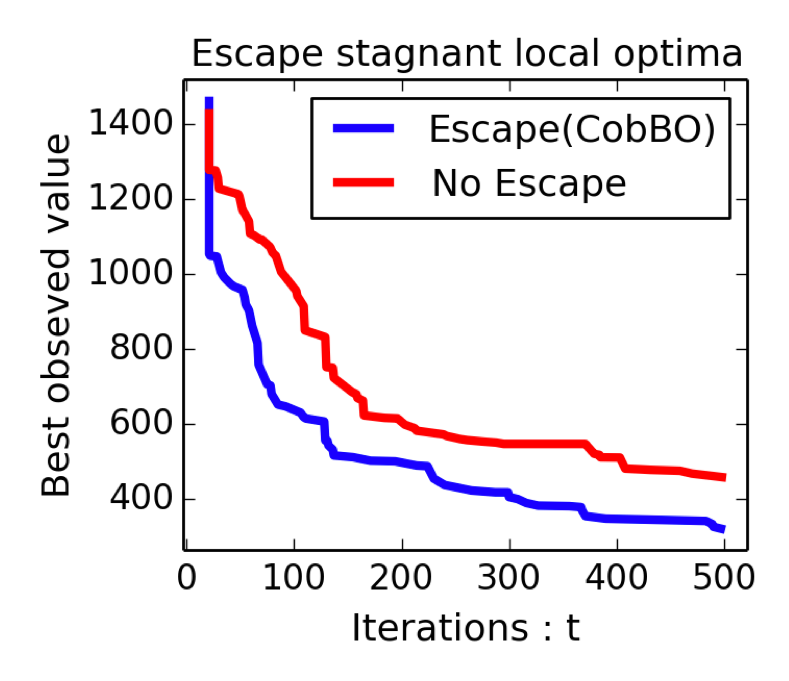}
    \caption{Ablation study for escaping local optima for Rastrigin on $[-5,10]^{50}$ with $20$ initial random samples. }
    %The results from 10 runs for each configuration are presented.}
  \label{fig:escape_ablation}
\end{figure}

CobBO can be viewed as a variant of block coordinate ascent.
Each subspace $\Omega_t$ contains a pivot point $V_t$.
If fixing the coordinates' values incorrectly, one is condemned to move in a suboptimal subspace. Considering that those are determined by $V_{t}$, it has to be changed in the face of many consecutive failures to improve over $M_{t}$ in order to escape this trapped local maxima.
We do that by decreasing the observed function value at $V_{t}$ and setting $V_{t+1}$ as a selected sub-optimal random point in $\mathcal{X}_t$. Specifically, we randomly sample a few points (e.g., $5$) in $\mathcal{X}_t$ with their values above the median and pick the one furthest away from $V_{t}$.
Figure~\ref{fig:escape_ablation} shows that the way CobBO escapes local optima is beneficial.

% \begin{wrapfigure}{r}{0.5\textwidth}
%   \begin{center}
%       \includegraphics[width=0.98
%   \linewidth,height=!]{figures/escape.png}
%     \caption{Ablation study for escaping local optima for Rastrigin on $[-5,10]^{50}$ with $20$ initial random samples. }
%     %The best performing run out of 5 runs for each configuration is presented.}
%   \label{fig:escape_ablation}
%   \end{center}
% \end{wrapfigure}

We further the experiment with Levy and Ackley functions of 100 dimensions, as described in Section~\ref{ss:highD} to compute the fraction of queries that improve
 the already observed maximal points due to the change of~$V_t$.
 
 \begin{table}[h!]
\centering
\begin{tabular}{ |c|c|c| } 
\hline
Problem & Average \# improved queries & Average \# improved queries due to escaping\\
\hline \hline
Ackley & 228 & 15.3 \\
\hline
Levy & 155 & 3\\
\hline
\end{tabular}
\caption{The number of improved queries due to escaping local maxima}
\label{table:escaping}
\end{table}

We observe that optimizing the Levy function yields very few queries that improve the maximal points by changing the pivot point, while optimizing the Ackley function can benefit more from that.  

\section{Additional experiments}
\begin{figure}[!htb]
  \centering
    \includegraphics[width=0.6\textwidth]{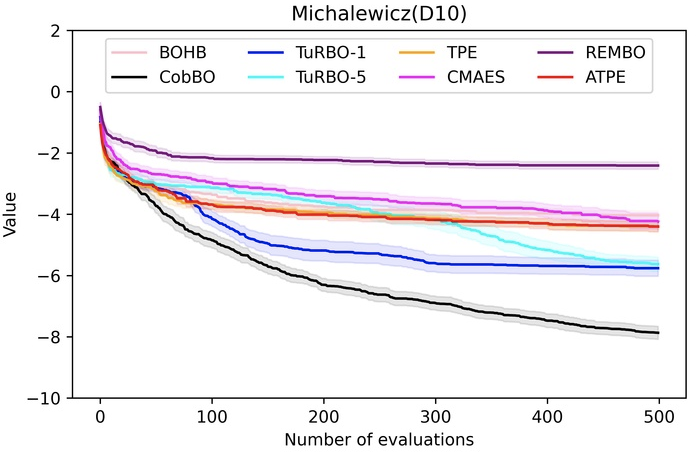}
     \caption{Performance over the low dimensional Michalewicz function with symmetrical and asymmetrical subspaces} 
    \label{fig:micha}
\end{figure}

%\subsection{Small-sized synthetic black-box functions (minimization)}
In Fig.~\ref{fig:micha} we show that CobBO also optimizes well the Michalewicz function on $10$ dimensions, although it has symmetric bumps, where certain subspaces pass through a point in a symmetrical manner and others break it. 
Other real applications include parameter tuning for recommendation systems, database online performance tuning, and simulation based parameter optimization. However, due to deviating from the main study of this paper, we refrain from presenting these results that require elaborated description on the application backgrounds.

\end{document}